\documentclass[10pt,twocolumn,letterpaper]{article}

\usepackage{cvpr}              % To produce the CAMERA-READY version
\usepackage[dvipsnames]{xcolor}
\usepackage{soul} % to highlight
\usepackage{lipsum}
\usepackage{xcolor}
\definecolor{lightblue}{rgb}{0.8,0.9,1}
\definecolor{lightgreen}{rgb}{0.9,1,0.9}
\definecolor{lightred}{rgb}{1,0.5,0.5}
\definecolor{gray}{rgb}{0.7,0.7,0.7}

\usepackage{graphicx}
\usepackage{amsmath}
\usepackage{amssymb}  % assumes amsmath package installed
\usepackage{lipsum}
\usepackage{multirow}

\usepackage{amsthm}
\newtheorem{definition}{Definition}
\newtheorem{thm}{Theorem}

\newtheorem{lemma}[thm]{Lemma}
\usepackage[ruled]{algorithm}
\usepackage{algpseudocode}

\usepackage{./rpm_SIunits}
\usepackage{./rpm_acronyms}
\usepackage{./rpm_math}
\usepackage{./rpm_misc}

\definecolor{cvprblue}{rgb}{0.21,0.49,0.74}
\usepackage[pagebackref,breaklinks,colorlinks,citecolor=cvprblue]{hyperref}
\def\paperID{9619} % *** Enter the Paper ID here
\def\confName{CVPR}
\def\confYear{2024}

\title{Unbiased Estimator for Distorted Conics in Camera Calibration}

\author{Chaehyeon Song, Jaeho Shin, Myung-Hwan Jeon, Jongwoo Lim, Ayoung Kim\\
Seoul National University\\
{\tt\small \{chaehyeon, leah100, myunghwan.jeon, jongwoo.lim, ayoungk\}@snu.ac.kr}
}

\begin{document}
\maketitle
\begin{abstract}
In the literature, points and conics have been major features for camera geometric calibration. 
Although conics are more informative features than points, the loss of the conic property under distortion has critically limited the utility of conic features in camera calibration. Many existing approaches addressed conic-based calibration by ignoring distortion or introducing 3D spherical targets to circumvent this limitation. In this paper, we present a novel formulation for conic-based calibration using moments. Our derivation is based on the mathematical finding that the first moment can be estimated without bias even under distortion. This allows us to track moment changes during projection and distortion, ensuring the preservation of the first moment of the distorted conic. With an unbiased estimator, the circular patterns can be accurately detected at the sub-pixel level and can now be fully exploited for an entire calibration pipeline, resulting in significantly improved calibration.
The entire code is readily available from 
\small \url{https://github.com/ChaehyeonSong/discocal}.
\normalsize
\end{abstract}
    
\section{Introduction}
\label{sec:intro}

\begin{figure}[!t]
    \centering
    \includegraphics[trim=20 20 30 30, clip,width=0.95\columnwidth]{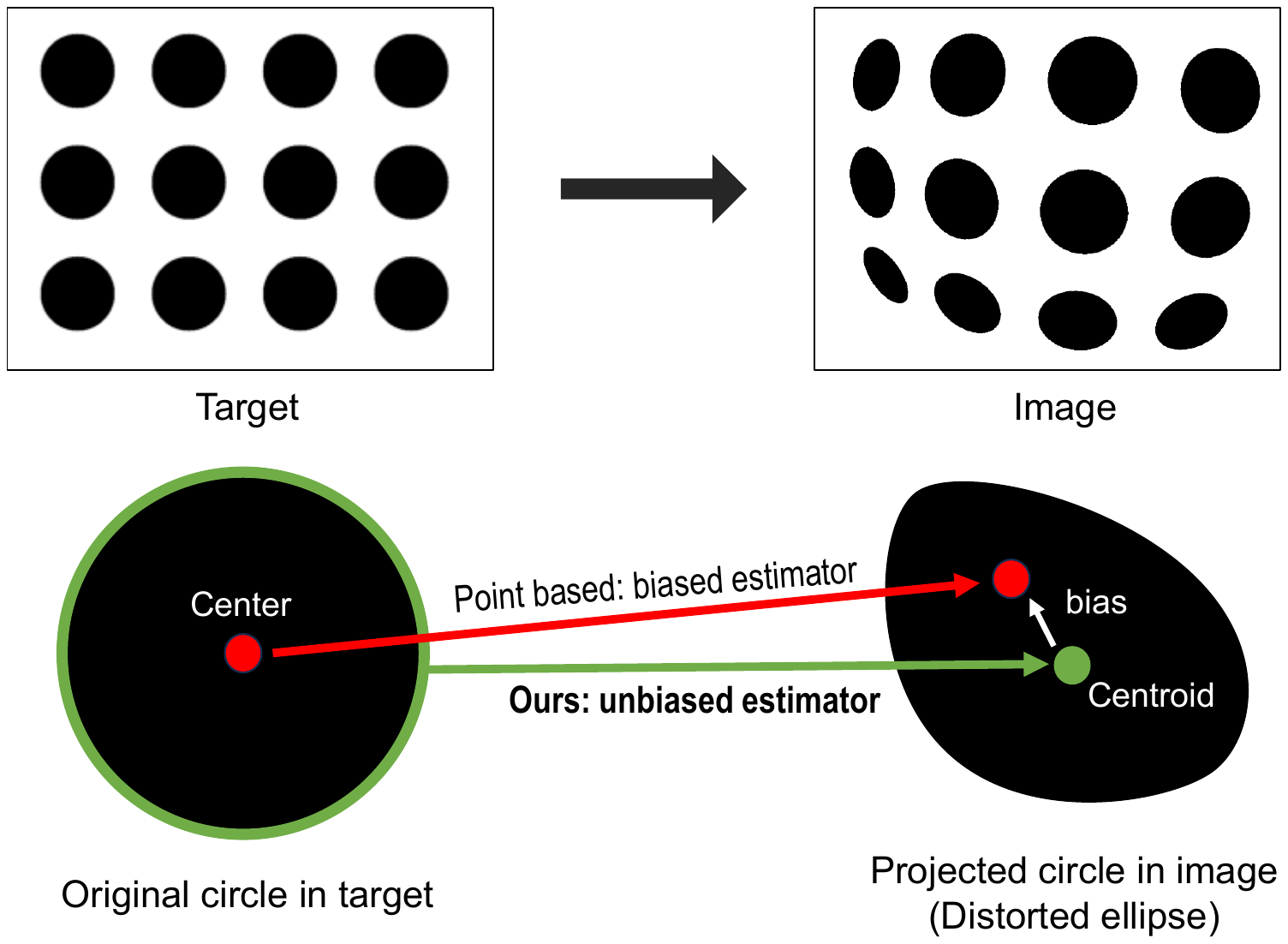}
    \caption{\textbf{Centroid estimation in distorted Images.} This figure illustrates the effects of camera projection and lens distortion on circular targets within an image. Due to these distortions, circles lose their conic properties and deform into distorted ellipses, making center point tracking challenging. (Red) failure of conventional control point estimation methods, as indicated by an incorrectly tracked center point. (Green) the proposed unbiased estimator accurately identifies the center point of the transformed ellipse, as derived from closed-form calculations.}
    \label{fig:overview}
    \vspace{-5mm}
\end{figure}

Camera calibration is essential in 3D computer vision and vision-based perception. Significantly, the processes of understanding the geometry from images rely on accurate camera calibration, including 3D dense reconstruction \cite{ECCV-2020-mildenhall, ToG-2022-muller}, visual \ac{SLAM}~\cite{TRO-2015-mur, TRO-2018-qin, RAL-2022-jeon}, and depth estimation~\cite{CVPR-2017-godard}. This fundamental problem has been tackled in various ways~\cite{ECCV-2004-chen, PAMI-2005-kim,IP-2010-wong, OL-2015-sun, IP-2021-chuang} and the most calibration methods exploit planar targets covered with particular patterns such as grid structure of squares~\cite{PAMI-2000-zhang, CVPR-1999-sturm} or circles~\cite{PAMI-2000-heikkila, PAMI-2006-kannala}. This planar target-based approach requires precise measurements with an unbiased projection model of control points (i.e., corner of the square or centroid of the circle) to achieve accurate calibration results. Much literature has studied the distinct advantages and disadvantages of the two patterns. The checkerboard pattern ensures precise (unbiased) estimation in projective transformation and distortion, yet it is limited to pixel-level detection accuracy. On the other hand, the circular pattern~\cite{ICPR-1996-heikkila} excels in achieving sub-pixel level detection accuracy, while the biased projection model has led to poor calibration results. Compensating the perspective bias using the conic feature is not sufficient, and the more dominant factor is the nonlinear lens distortion~\cite{PAL-2007-mallon}. It degrades the geometric properties of the cone, making it difficult to estimate the centroid of the projected circle as described in \cref{fig:overview}.

%FIGURE
%FIGURE

In this paper, we push the envelope on the circular pattern by proposing an unbiased estimator in a closed-form solution to handle the distortion bias. To the best of our knowledge, this is the first analytic solution describing projected conic features under radial polynomial distortion. Inspired by the fact that moment representation is possible to describe the general distribution transformation under any polynomial mapping, we proved that the characteristics of the distorted conic, such as the centroid, could be expressed as a linear combination of $n^\text{th}$ moments of the undistorted conic. Our work is not only limited to distortion models for pinhole camera calibration. Moreover, we suggest a general and differentiable approach to track the conic under arbitrary nonlinear transformations that can be approximated as polynomial functions. 

By conducting both synthetic and real experiments, we have validated that our estimator is unbiased by investigating the reprojection error under various distortions and circle radius with known camera parameters. Our estimator is also applied to the calibration of RGB and \ac{TIR} cameras which have difficulty in detecting control points due to the boundary blur effect (see \bl{\apref{Appendix:TIR_Char}}). As a result, our method outperforms existing circular pattern-based methods and the checkerboard method in reprojection error and 6D pose manner. The main contribution of our work is summarized as follows.

\begin{itemize}
    \item We pioneer the unbiased estimator for distorted conic to fully exploit the virtue of circular patterns inheriting a simple, robust, and accurate detector. {In a blend of mathematical elegance and practical ingenuity, our work completes the missing piece in the conic-based calibration pipeline.} 
    \item 
    We leverage the probabilistic concept of a moment, which had not been previously attempted in the calibration, and provide general moments of conic as an analytic form with thorough mathematical derivation and proof.
    This approach enables us to design the unbiased estimator for circular patterns.
    \item Our unbiased estimator improves the overall calibration performance when tested on both synthetic and real images. Especially, we showcase that our method yields substantially improved calibration results for \ac{TIR} images, which often include high levels of blur, noise, and significant distortion. 
    We open our algorithm to support the community using the conic features in their calibration. 

\end{itemize}

\section{Related Works}
\label{sec:related_work}

\textbf{Planar pattern and control point.}
\citet{PAMI-2000-zhang} and \citet{CVPR-1999-sturm} introduced a calibration method utilizing a planar target with some specific pattern printed on it. The planar target should include control points that can be easily and accurately extracted from the specific pattern. The checkerboard pattern comprising black and white squares was initially considered, and then the alternative pattern of circles with grid structure was introduced by \citet{PAMI-2000-heikkila}. The checkerboard pattern uses the corners of squares as control points, and the circular pattern uses the centroid of projected circles in the image with subpixel accuracy \cite{ICPR-1996-heikkila}. Since the exact position of control points is crucial, an interactive way to refine the location of control points was also considered~\cite{ICCV-2009-datta}.

\noindent\textbf{Unbiased estimator.}
To achieve accurate calibration results, the quality of measurement and estimation value of the control point are both vital. In contrast to a checkerboard pattern, which is proven to have an unbiased estimator in most cases, applying the same estimator to a circular pattern results in bias~\cite{PAL-2007-mallon}. Specifically, the center point of circles in the target is not projected to the center point of the projected circle in the image. Several works~\cite{PAMI-2000-heikkila,PAMI-2005-kim} challenged this problem by adopting conic-based transformation. These methods achieved unbiased estimators under linear transformation, such as perspective transformation. However, bias resulting from distortion is unresolved since conic characteristics are not preserved under nonlinear transformation. The bias originating from distortion is a dominant factor in most cameras~\cite{PAL-2007-mallon}. To resolve this issue, \citet{PAMI-2006-kannala} introduced a generalized concept of the unbiased estimator for the circular pattern but could not derive an analytic solution for the given integral equation.

\noindent\textbf{Methods without control point.}
The approach to directly use the conic characteristic has been developed simultaneously. By utilizing concentric circles, early works~\cite{ECCV-2004-chen, PAMI-2005-kim} could obtain the image of absolute conic from the conic equations. Some works \cite{IP-2010-wong, OL-2015-sun} focused on the sphere, whose projected shape only depends on the configuration between the center point and the camera. 
Unfortunately, these methods also suffered from distortion bias, which destroys all crucial geometry of the conic or sphere. To deal with the nonlinear distortion function, \citet{MVA-2001-devernay} leveraged the line feature, which is always straight in undistorted images. Although this work did not explicitly estimate the intrinsic parameter, recent work~\cite{IP-2021-chuang} introduced the closed-form solution of the intrinsic parameter using line features with the undistorted images.

\noindent\textbf{General model approach.} The existing camera model assumes an ideal lens and approximates the nonlinear mapping with few parameters. Alleviating this assumption, a more general ray tracing model has been introduced. \citet{CVPR-2020-schops} suggested a general un-projection model using B-spline interpolation with nearest points. Another works~\cite{CVPR-2022-pan} proposed pixel-wise focal length to consider general nonlinear mapping at the radial direction. These alternative approaches possess potential; however, existing applications of 3D vision still need geometric camera models. Therefore, in this paper, we focused on geometric model-based calibration, especially the pin-hole model.
\section{Preliminary}
\label{sec:preliminary}

%-------------------------------------------------------------------------%
\subsection{Notations}

\begin{figure}[!t]
    \centering
    \includegraphics[trim=43 120 43 0, clip,width=0.99\columnwidth]{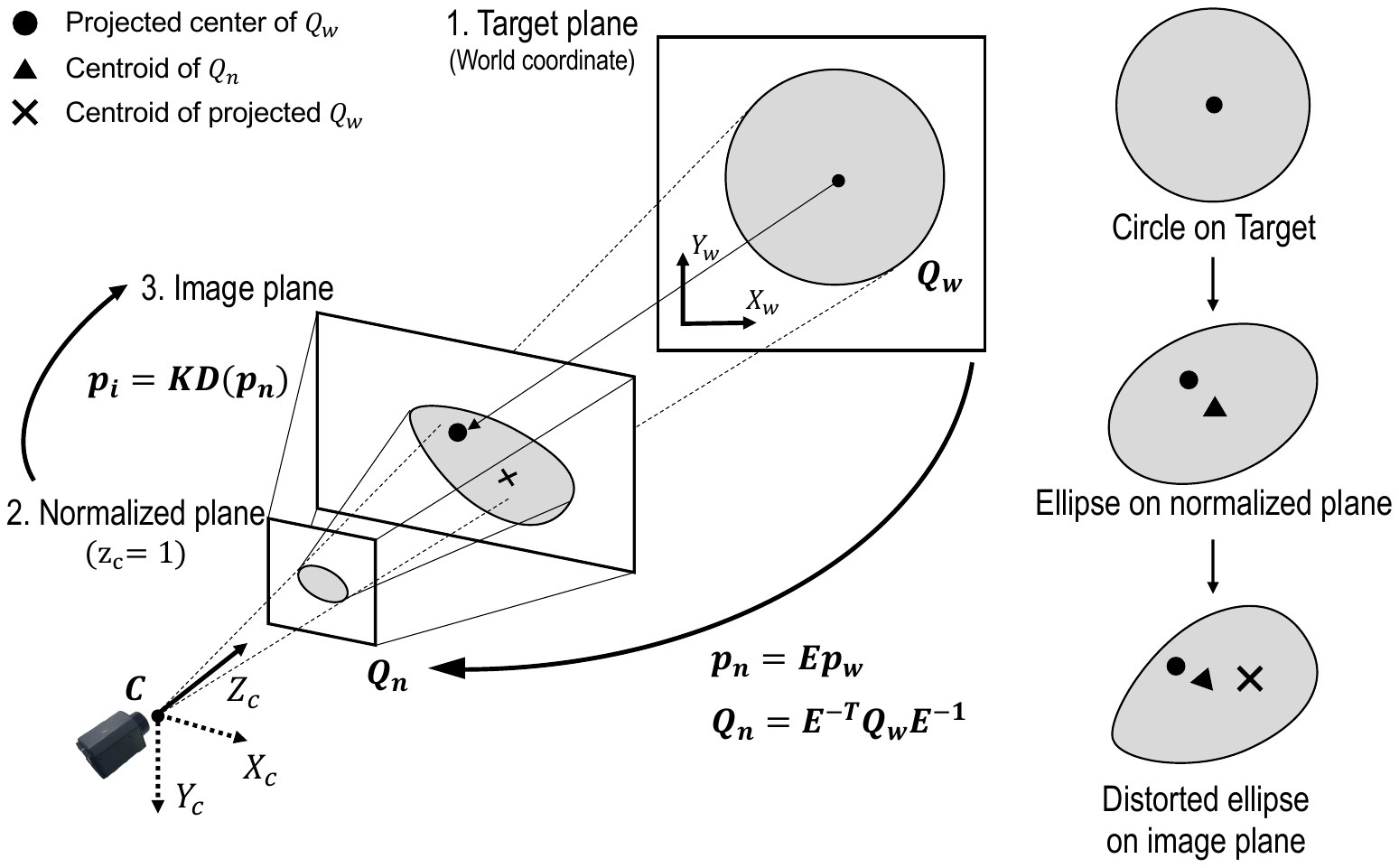}
    \caption{\textbf{Image projection geometry.} Due to projection and lens distortion, there exists some mismatch between the projected center of the circle on the target plane (circled dot), projected center of ellipse on the normalized plane(triangle sign), and the actual centroid of the shape on the image plane (crossed sign). The transformed shape, caused by non-linear distortion in the normalized plane, cannot be analytically described as the original conic. However, despite these distortions, our algorithm successfully locates the true centroid through moment tracking.}
    \label{fig:geometry}
    \vspace{-5mm}
\end{figure}

A vector $\bs{p} \in \Re^n$ and a matrix $\bs{Q} \in \Re^{n\times n}$ are denoted by lowercase and uppercase in bold with the coordinate in the subscript. We set the target plane to be the same as the $xy$-plane of the world coordinate. Thus, $\bs{p}_w=(x_w,y_w)^\top$ is a 2D point in the target plane written in the target coordinate with $\bs{p}_i$ being a corresponding point in the image. This target point $\bs{p}_w=(x_w,y_w)^\top$ is projected to a point $\bs{p}_n$ in the normalized plane via perspective projection, and then $\bs{p}_n$ is mapped to a point $\bs{p}_i$ in image plane under distortion and intrinsic matrix.
%$\bs{p}_n$ is a 2D point in the normalized plane at the camera coordinate and $\bs{p}_d$ is the distorted point in the normalized plane corresponding to $\bs{p}_n$. 
Also, $\bs{Q}_w$ is a matrix representation of an ellipse in the target plane, and $\bs{Q}_n$ is an ellipse in the normalized plane. More details of these notations and the geometric relationship between the coordinates are described in \figref{fig:geometry}. We use $\sim$ for a homogeneous vector representation. Also, $\Tilde{\bs{p}} \simeq  \Tilde{\bs{q}}$ means that two vectors are identical up-to-scale.

\subsection{Matrix representation of conic}

A conic is a set of points $(x,y)$ that satisfy the following equation:
\begin{equation}
    ax^2 + 2bxy+cy^2 +2dx + 2ey + f =0,
\end{equation}
which can be also written in matrix form as:
\begin{equation}
    \bs{x}^{\top}\bs{Q}\bs{x} = 0, \hspace{4mm}
    \bs{Q} = \begin{pmatrix}
        a & b & d\\
        b & c & e\\
        d & e & f
    \end{pmatrix}, \hspace{2mm}
    \bs{x} = \begin{bmatrix}
        x\\ y \\ 1  
    \end{bmatrix}.
\end{equation}
The symmetric matrix $\bs{Q}$ is the characteristic matrix of the conic. In general, the conic comprises an ellipse, parabola, and hyperbola, but we only concentrate on the ellipse in this paper since the images of circles are ellipses except for extreme cases. Given characteristic matrix $\bs{Q}$, the center of an ellipse is calculated as:
\begin{equation}
    \label{eq:conic8}
    \Tilde{\bs{p}} = \bs{Q}^{-1} \begin{bmatrix}
        0 & 0 & 1
    \end{bmatrix}^\top.
    % \Tilde{\bs{p}} = \begin{bmatrix}
    %     x \\ y \\ 1 
    % \end{bmatrix}
    % = \bs{Q}^{-1}\begin{bmatrix}
    %     0 \\ 0 \\ 1
    % \end{bmatrix}.
\end{equation} Detailed derivation for geometric features of the ellipse is explained in \bl{\apref{Appendix:Ellipse_feature}}. 

%-------------------------------------------------------------------------% 

\subsection{Camera model}

Adopting a pinhole camera model, a 2D point $\Tilde{\bs{p}}_w$ in the target plane is projected to the image by:
\begin{eqnarray}
  \bs{K} &=& \begin{bmatrix}
        f_x & \eta & c_x\\
        0 & f_y & c_y\\
        0 & 0 & 1
    \end{bmatrix},\\
  \bs{T}_{cw} &=& \begin{bmatrix}
        \bs{R} & \bs{t}\\
        \bs{0}^{\top} & 1
    \end{bmatrix} = \begin{bmatrix}
        \bs{r}_1 & \bs{r}_2 & \bs{r}_3& \bs{t}\\
        0 & 0 & 0 & 1
  \end{bmatrix}, \\
  \Tilde{\bs{p}}_i &=&\begin{bmatrix}
        u\\v\\1
    \end{bmatrix}
    \simeq \bs{K}\begin{bmatrix}
        \bs{r}_1 & \bs{r}_2 & \bs{r}_3&\bs{t}\\
    \end{bmatrix}\begin{bmatrix}
        x_w\\
        y_w\\
        z_w\\
        1
  \end{bmatrix} \\
   & \simeq & \bs{K} \begin{bmatrix}
        \bs{r}_1 & \bs{r}_2 & \bs{t}\\
    \end{bmatrix}\begin{bmatrix}
        x_w\\
        y_w\\
        1
    \end{bmatrix}\hspace{2mm} (\because z_w=0)\\
        & \simeq & \bs{K} \bs{E} \Tilde{\bs{p}}_w \hspace{3mm}(\Tilde{\bs{p}}_w \triangleq [x_w, y_w, 1]^{\top})\\
    & \simeq & \bs{K}\Tilde{\bs{p}}_n \simeq \bs{H}\Tilde{\bs{p}}_w .
\end{eqnarray}  % TO MANUALLY BREAK INTO PAGES  

\noindent Here, $\bs{K}$ is the intrinsic matrix of the camera, $\bs{E}$ is the extrinsic parameter between the target coordinate and the camera coordinate, and $\bs{H}$ is the homography matrix. The $z_w$ is zero since the $p_w$ is a point located on the $xy$-plane. The above equation fully determines the relation between the target point and the projected point ($\Tilde{\bs{p}}_n$) in the normalized plane.
Unfortunately, this projected point ($\Tilde{\bs{p}}_n$) further undergoes a nonlinear transformation (lens distortion) when projected onto the image plane. Between the two types of distortion, radial and tangential, we only consider radial distortion, which is known to be sufficient in most cameras \cite{CVPR-2019-lopez,CVPR-2022-pan}. The radial distortion is typically modeled with a polynomial function \cite{PE-1996-brown, MNRAS-1919-conrady} as:
\begin{eqnarray}
    \label{eq:sn}
    s_n &=& x_n^2 + y_n^2,\\
    \label{eq:def_dist}
    k &=& \sum_{i=0}^{n_d}d_is_n^i \hspace{5mm} (d_0=1),\\
    \Tilde{\bs{p}}_d &=& D(\Tilde{\bs{p}}_n)=[x_d, y_d, 1]^\top=[kx_n, ky_n, 1]^\top,\\
    \Tilde{\bs{p}}_i &\simeq& \bs{K}\Tilde{\bs{p}}_d = \bs{K}D(\Tilde{\bs{p}}_n) = \bs{K}D(\bs{E}\Tilde{\bs{p}}_w),
\end{eqnarray}
where $d_i$ are the distortion parameters and $D$ is the distortion function. The value $n_d$ is the maximum order of distortion parameters, typically less than four in existing calibration methods~\cite{matlab-2010-bouguet}.

\subsection{Measurement model (Control point)}
The centroid of a black dot in an image is obtained as
\begin{eqnarray}
    \Bar{\bs{p}}_i = \frac{\int w\bs{p}_i dA_i}{\int wdA_i}
    \approx \frac{\sum_{j} w_j\bs{p}_i(j)}{\sum_j w_j} 
\end{eqnarray}
In the ideal case, $w$ is $(255-\text{intensity})$. In the real world, other light sources frequently distort the color value of the projected circle so that $w$ is set to one for robustness.

\section{Unbaised Estimator for Circular Patterns}
\label{sec:method}

%-------------------------------------------------------------------------%    
\subsection{Intuition: moment approach}
\label{sec:momentum}

The main objective of this section is to provide a thorough mathematical derivation of the unbiased estimator of the control point, which is the center point of the projected circle in the image originating from the target pattern. Tracking the control point under homography transformation is straightforward. An ellipse is projected to another ellipse under homography transformation $\bs{H}$ as: 
\begin{eqnarray}
    \Tilde{\bs{p}}_f &\simeq& \bs{H}\Tilde{\bs{p}}_i,\\
    \bs{Q}_f & \simeq &\bs{H}^{-\top}\bs{Q}_i\bs{H}^{-1}. \label{eq:1_2}
\end{eqnarray}
The subscript indicates the original frame $i$ to transformed frame $f$. By combining Eqs. \eqref{eq:conic8} and \eqref{eq:1_2}, the center of the transformed ellipse is calculated as $\bs{H}\bs{Q}_i^{-1}\bs{H}^\top(0,0,1)^\top$ and it is not the same as the transformed center point of the original ellipse, $\bs{H}\bs{Q}_i^{-1}(0,0,1)^\top$. Before adapting our approach, we define some concepts of geometry as preliminary.

\begin{definition}[Shape]
    A shape $A_{xy}$ is a set of points enclosed by the closed curve in xy-plane. $|A_{xy}|$ is the inner area of the closed curve. $A_k$ denotes $A_{x_ky_k}$.
\end{definition}

The conic feature suffers from losing its characteristic under nonlinear transformations like distortion. To build an unbiased estimator under distortion, we utilize moment theory. In probability theory, $(i+j)^{\text{th}}$ moment of random variables $X$ and $Y$ is defined as $E[X^iY^j]$. By assuming points of the given shape lie on uniform distribution, the spatial average corresponds to the expectation of a random variable. We define the $(i+j)^{\text{th}}$ moment of a 2D shape in $xy$ coordinate.

\begin{definition}[Moment]
    For any $m,n \in \mathbb{Z}^*$(nonnegative integer set), $M^{m,n}_{xy}$ is $(m+n)^{\text{th}}$ moment of $A_{xy}$ such that
    \begin{equation}
        M^{m,n}_{xy} \triangleq \frac{1}{|A_{xy}|} \int x^my^n dA_{xy}.
    \end{equation}
\end{definition}

The first-moment factors ($M^{1,0}$ and $M^{0,1}$) are the center point of the shape, and the second-moment factors ($M^{2,0}$, $M^{1,1}$, and $M^{0,2}$) are related to the covariance of the shape. Therefore, an ellipse can be described by only using the first and second-moment factors. The matrix form of ellipse is $\bs{x}^\top\bs{Q}\bs{x} \le 0$, and the moment form is $[M^{1,0},M^{0,1},M^{2,0},M^{1,1},M^{0,1}]$. Each representation has the same five-\ac{DoF}.

%THOEREM
\begin{thm}
\label{thm:1}
    Given a polynomial function $D : (x,y) \longrightarrow (x',y')$ which is invertible in domain $X$, let $A_{xy} \subset X$ is transformed to $A_{x'y'}$ by $D$. For any $m, n \in \mathbb{Z^*}$, there exist $c_{ij}(D)\in \mathbb{R}$ and $p, q \in \mathbb{Z}^*$ such that
    \begin{equation*}
        |A_{x'y'}|M_{x'y'}^{m,n} = |A_{xy}|\sum_{i=0}^{p}\sum_{j=0}^{q} c_{ij} M_{xy}^{i,j}.
    \end{equation*}
\end{thm}
\begin{proof}
    See \bl{\apref{Appendix:Theorem1}}. 
\end{proof}

% \begin{proof}
% \small
% \begin{eqnarray}
%   [x',y'] &=& D(x,y) = [f(x,y), g(x,y)]\\
%   \bs{J} &=& Jacob(D) = \begin{bmatrix}
%         \frac{\partial f}{\partial x} && \frac{\partial f}{\partial y}\\
%         \frac{\partial g}{\partial x} && \frac{\partial g}{\partial y}
%         \end{bmatrix},\\
%   |A_{x'y'}|M_f^{nm} &=& \int (x')^n(y')^m dA_f \label{eq:momentum1}\\
%     &=& \int f(x,y)^n g(x,y)^m \det(\bs{J})dA_{xy} \label{eq:Mnm}
% \end{eqnarray}
% %
% Since $f(x,y)$, $g(x,y)$, and $det(\bs{J})$ are polynomial of $x$ and $y$,
% %
% \begin{eqnarray}
%   |A_{x'y'}|M_f^{nm}  &=& |A_{xy}|\left[ \frac{1}{|A_{xy}|} \int \sum_{ij}c_{ij} x^iy^jdA_{xy} \right]\\
%   &=&  |A_{xy}| \sum_{ij}c_{ij} \left[ \frac{1}{|A_{xy}|} \int x^iy^jdA_{xy} \right] \\
%   &=& |A_{xy}|\sum_{i=0}^{p}\sum_{j=0}^{q} c_{ij}M_{xy}^{ij}.
% \end{eqnarray}
% \normalsize
% \end{proof}
%THOEREM
%
% \vspace{-0.4cm}
\thmref{thm:1} implies that for any polynomial mapping, the moment of the transformed shape can always be expressed by the linear combination of the moment of the original shape. The required number of $p$ and $q$ depends on the order of the polynomial function.

%-------------------------------------------------------------------------%    
\subsection{Tracking control point under distortion}
\label{sec:tracking}
Even though it is impossible to describe a distorted ellipse in any analytic form, its moment representation always exists by \thmref{thm:1}. Despite using high-order moments, we only use the first moments, which is the centroid of shape. The reason is explained in the next section \cref{sec:reason1}. 

To examine the shape under distortion mapping, let $A_n$ be a shape in a normalized plane comprising points $(x_n, y_n)$ and $A_d$ be the distorted shape comprising points $(x_d, y_d)$. Using \thmref{thm:1}, the center of the distorted ellipse in the normalized plane can be calculated as follows.
\small
\begin{eqnarray}
    w_{0r} &=& \sum_{i}(2i+1)d_id_{r-i} \\
    w_{1r} &=& \sum_{i}\sum_{j}  (2i+1)d_id_jd_{r-i-j}\\
    \label{eq:M00}
  1 &=& M_d^{0,0} = \frac{|A_n|}{|A_d|}\sum_{r=0}^{2n_d} w_{0r} \left[ \frac{1}{|A_n|} \int s_n^r dA_n\right] \\
  \label{eq:M10}
  \Bar{x}_d &=& M_d^{1,0} = \frac{|A_n|}{|A_d|} \sum_{r=0}^{3n_d} w_{1r} \left[ \frac{1}{|A_n|} \int x_ns_n^r dA_n\right] \\
  \label{eq:M01}
  \Bar{y}_d &=& M_d^{0,1} = \frac{|A_n|}{|A_d|} \sum_{r=0}^{3n_d} w_{1r} \left[ \frac{1}{|A_n|} \int y_ns_n^r dA_n\right]
\end{eqnarray}
\normalsize
Refer to \equref{eq:sn} for the symbol $s_n$ defined as $x_n^2+y_n^2$ and \equref{eq:M00} is need for calculate $\frac{|A_n|}{|A_d|}$.
The $2n_d$ and $3n_d$ come from calculating the Riemannian Metric and the details are provided in the \bl{\apref{Appendix:Moment_tracking}}.

The centroid of the projected shape in the image plane is readily calculated as below with details in \bl{\apref{Appendix:ImageCenterPoint}}.
\begin{figure}[t]
    \centering
    \includegraphics[trim=40 80 60 75, clip,width=0.99\columnwidth]{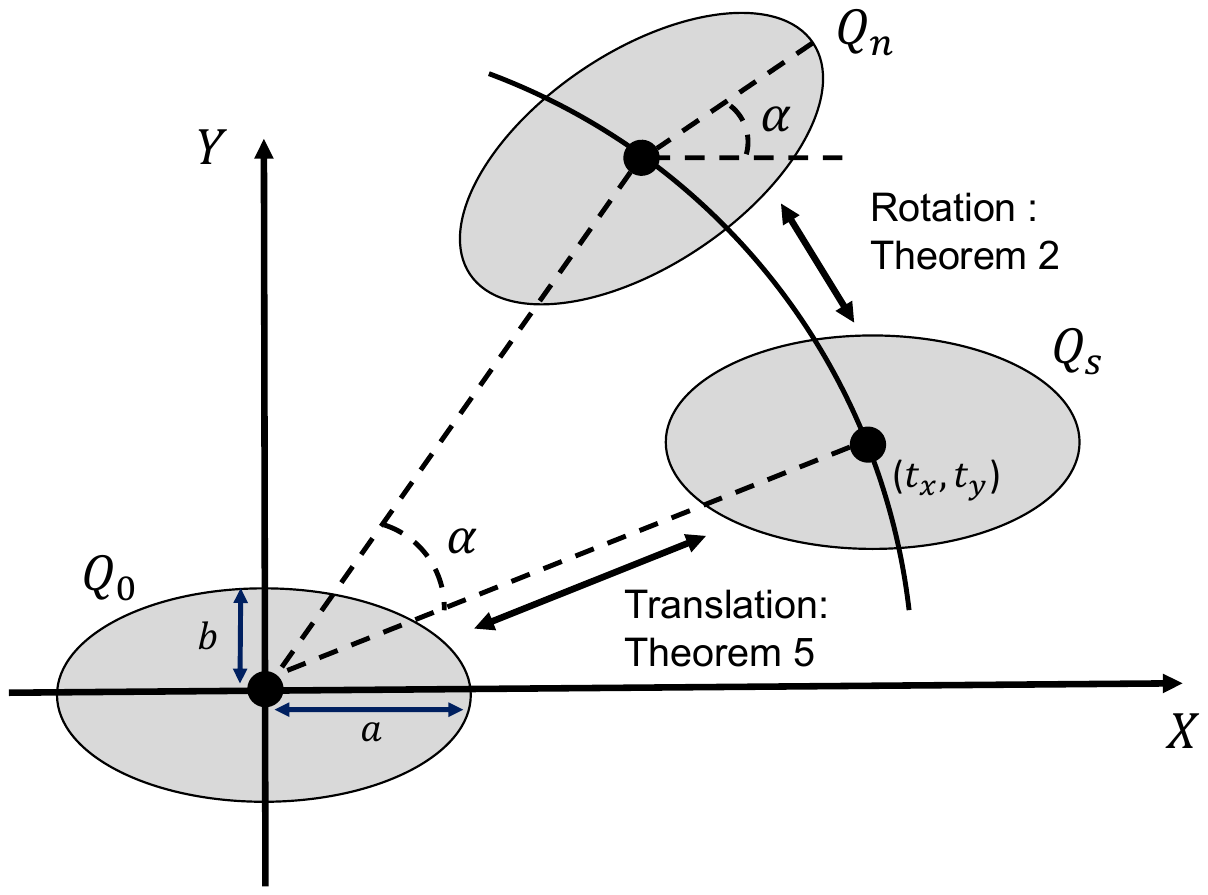}
    \caption{\textbf{Moment calculation strategy for a arbitrary ellipse $\boldsymbol{Q}_n$.} 
    The moments of the rotated ellipse $\boldsymbol{Q}_n$ are obtained from the moments of the un-rotated standard ellipse $\boldsymbol{Q}_s$ using \thmref{thm:rot}. The moments of $\boldsymbol{Q}_s$ are obtained from moments of another standard ellipse $\boldsymbol{Q}_0$, which is located at the origin, using \thmref{thm:vr}.}

    \label{fig:moment_geo}
    \vspace{-5mm}
\end{figure}
\begin{eqnarray}
    \Bar{x}_i &=& f_x\Bar{x}_d +\eta \Bar{y}_d + c_x\\
    \Bar{y}_i &=& f_y\Bar{y}_d + c_y.
\end{eqnarray}

\noindent Therefore, to build the unbiased estimator of $[\Bar{x}_i, \Bar{y}_i]$, we only need $\frac{1}{|A_n|} \int s_n^r dA_n$, $ \frac{1}{|A_n|} \int x_ns_n^r dA_n$, and $ \frac{1}{|A_n|} \int y_ns_n^r dA_n$ for every integer $r$ from $0$ to $3n_d$, where $(x_n, y_n)$ is a point in the set $A_n = \{\bs{p}_n|\, \Tilde{\bs{p}}_n^T\bs{Q}_n\Tilde{\bs{p}}_n \le 0\}$. We define these values as a vector $\bs{v}_n^r$:
\begin{equation}
    \bs{v}_n^r \triangleq \begin{bmatrix}
        \frac{1}{|A_n|} \int  x_ns_n^rd A_n \\
        \frac{1}{|A_n|} \int y_ns_n^rd A_n \\
        \frac{1}{|A_n|} \int s_n^rd A_n 
    \end{bmatrix}
\end{equation}
Calculating $\bs{v}_n^r$ directly from an ellipse on a normalized plane is not straightforward; therefore, we divide the process into two steps as shown in \cref{fig:moment_geo}.

%THEOREM
\begin{thm}[Rotation Equivariant]
\label{thm:rot}
For any 2D rotation transformation $R : (x_s, y_s) \rightarrow (x_n, y_n)$, there exist $\alpha\in[0, 2\pi]$ such that, 
\begin{eqnarray}
    x_n &=& \cos\alpha x_s -\sin\alpha y_s \nonumber\\
    y_n &=& \sin\alpha x_s + \cos\alpha y_s \nonumber\\
    \bs{v}_n^r &=&
    \begin{bmatrix}
        \cos\alpha & -\sin\alpha &0\\
        \sin\alpha & \cos\alpha & 0\\
        0 & 0 & 1\\
    \end{bmatrix} 
    \bs{v}_s^r \label{eq:vnr}
\end{eqnarray}
\end{thm}
\begin{proof}
    See \bl{\apref{Appendix:Theorem2}
}. 
\end{proof}

\noindent \thmref{thm:rot} implies that it is possible to obtain $\bs{v}_n^r$ of arbitrary ellipse $A_n$ if we can calculate the vector $\bs{v}_s^r$ of any unrotated ellipse $A_s$ whose major and minor axis are parallel to the axis of the coordinate system. 
%LEMMA
\begin{lemma}
\label{lemma:1}
Consider $I^{m,n}= \frac{1}{2\pi}\int_0^{2\pi} \cos^{m}\theta\sin^n\theta d\theta$ with $m,n,i,j \in \mathbb{Z}^*$ then,
\begin{equation*}
    I^{m,n}= \begin{cases}
        \frac{\binom{2i+2j}{i+j}\binom{i+j}{i}}{\binom{2i+2j}{2i}2^{2i+2j}} & \text{if } m = 2i, n = 2j\\
        0 & otherwise.
    \end{cases}
\end{equation*}
\end{lemma}
\begin{proof}
    See \bl{\apref{Appendix:Lemma3}}. 
\end{proof}

\vspace{0.05cm}

\begin{lemma}
\label{lemma:2}
The $(m+n)^{th}$ moment of $A_{0}=\{(x_0,y_0)|(x_0/a)^2 +(y_0/b)^2 \le 1\}$ is
\begin{equation*}
     M^{m,n}_0= \frac{ a^mb^n}{1+(m+n)/2}I^{m,n}
\end{equation*}
\end{lemma}
\begin{proof}
    See \bl{\apref{Appendix:Lemma4}}.
\end{proof}
\vspace{0.05cm}

We can obtain $\bs{v}_s^r$ from \thmref{thm:vr} using \lemref{lemma:1} and \lemref{lemma:2}. $M_0^{m,n}$ denotes analytic solutions for $(m+n)^{\text{th}}$ moments of the unrotated ellipse located at the origin.

\begin{thm}[Solution of $\bs{v}_s^r$]
\label{thm:vr}
When $x_s$ and $y_s$ satisfy $((x_s-t_x)/a)^2 +((y_s-t_y)/b)^2 \le 1$ then,

\scriptsize
    \begin{align*}
        \bs{v}_s^r[0] &=\sum_{i=0}^{r}\sum_{j=0}^{r-i} M_0^{2i,2j} \sum_{k=i}^{r-j} \binom{r}{k} \binom{2k+1}{2i}\binom{2r-2k}{2j}t_x^{2k-2i+1}t_y^{2r-2k-2j}\\
        \bs{v}_s^r[1] &=\sum_{i=0}^{r}\sum_{j=0}^{r-i} M_0^{2i,2j} \sum_{k=i}^{r-j} \binom{n}{k} \binom{2k}{2i}\binom{2r-2k+1}{2j}t_x^{2k-2i}t_y^{2r-2k-2j+1}\\
        \bs{v}_s^r[2] &=\sum_{i=0}^{r}\sum_{j=0}^{r-i} M_0^{2i,2j} \sum_{k=i}^{r-j} \binom{r}{k} \binom{2k}{2i}\binom{2r-2k}{2j}t_x^{2k-2i}t_y^{2r-2k-2j}
    \end{align*}
\normalsize
\end{thm}
\begin{proof}
    See \bl{ \apref{Appendix:Theorem5}}.
\end{proof}

We have reduced the computational cost of calculating the $\bs{v}_s^r$ using dynamic programming. The coefficients comprise products of binomials while $I^{m,n}$ is invariant to the conic shape; therefore, it is possible to calculate and store these values in advance. Hence, $\bs{v}_s^r$ is obtained in $O(1)$\footnote{Indeed, $O(n_d^3)$. However, $n_d$ is a constant number and typically lower than four so that we can treat it as $O(1)$}. The $\bs{v}^r_n$ can be obtained simply by multiplying the rotation matrix to $\bs{v}^r_s$ as \thmref{thm:rot}.
Using each of the previous derivations to compute the Eqs. \eqref{eq:M00}~to~\eqref{eq:M01}, the overall process of the unbiased estimator is described as \cref{alg:unbiased}.

%\begin{figure}[hb!]
\begin{algorithm}[h]
    \footnotesize
    \caption{Unbiased estimator}
    \label{alg:unbiased}
    \hspace*{\algorithmicindent} \textbf{Input:} \\% $\bs{Q}_w, \bs{E},\bs{K}, \bs{D}$\\
	\hspace*{\algorithmicindent}\hspace{1em} $\bs{Q}_w$ : matrix of a circle in target plane\\
	\hspace*{\algorithmicindent}\hspace{1em} $\bs{E} = [\bs{r}_1 \bs{r}_2 \bs{t}]$ : extrinsic parameter\\
	\hspace*{\algorithmicindent}\hspace{1em} $\bs{K}$ : intrinsic parameter\\
        \hspace*{\algorithmicindent}\hspace{1em} $\bs{D}=[1, d_1, d_2, d_3, ... ,d_n]$ : distortion parameter\\
	\hspace*{\algorithmicindent} \textbf{Output:} \\
        \hspace*{\algorithmicindent}\hspace{1em} $(x_i, y_i)$ : the center point of projected circle in image
    \begin{algorithmic}[1]

    \State $\bs{Q}_n = \bs{E}^{-\top}\bs{Q}_w\bs{E}^{-1}$
    \State $m_x, m_y, m_0=0$
    \For{$r = 0:3n_d$}
        \State $t_x', t_y', a, b, \alpha$ = \texttt{GeometryOfEllipse}($\bs{Q}_n$)\quad\text{\% \apref{Appendix:Ellipse_feature}}
        \State $t_x = \cos\alpha t_x' + \sin\alpha t_y'$ 
        \State $t_y = -\sin\alpha t_x' + \cos\alpha t_y'$ 
        \State $\bs{v}_s^r = \texttt{CalcualteVs}(t_x, t_y, a, b ,\bs{D},r)$\quad\text{\% \thmref{thm:vr}}
        \vspace{1mm}
        \State $\bs{v}_n^r = \bs{R}_{z}(\alpha)\bs{v}_n^r$\quad\text{ \% \thmref{thm:rot}}
        \State $w_{0r}, w_{1r} = \texttt{GetCoeff}(\bs{D},r)$\quad\text{\% \apref{Appendix:Moment_tracking}}
        \State $m_x = m_x + w_{1r}\bs{v}_n^r[0]$
        \State $m_y = m_y + w_{1r}\bs{v}_n^r[1]$
        \State $m_0 = m_0 + w_{0r}\bs{v}_n^r[2]$
    \EndFor
    \State $x_d = m_x/m_0$
    \State $y_d = m_y/m_0$
    \State $x_i = \bs{K}[0,0]x_d+\bs{K}[0,2]$
    \State $y_i = \bs{K}[1,1]y_d+\bs{K}[1,2]$
    \State \textbf{return} $x_i$, $y_i$
    \end{algorithmic}
\end{algorithm}

%-------------------------------------------------------------------------%  
% \vspace{-1cm}
\subsection{Robustness of the first moment}
\label{sec:reason1}
As mentioned at \cref{sec:momentum}, a conic is defined with second-order moments. Using the above results, it is possible to calculate the second-order moments of distorted conic in the image plane. However, the boundary blur effects easily contaminate high-order moments. For instance, if there is some dilation or erosion in the ellipse, the major and minor axis lengths become shorter or longer while the centroid of the shape is invariant. For calibration, the unbiased estimator and accurate measurement are both essential; therefore, utilizing only the first momentum is more beneficial for accurate calibration.
Another advantage of the first moment is its robustness to the image noise. Assume that there is some noise in the boundary points of the shape and the noise follows a normal distribution whose mean is zero and variance is $\sigma^2$, then the variance of the first moment of the shape is reduced by $1/n$. For boundary points following $X_i\sim \mathcal{N}(\mu, \sigma^2)$,
\begin{eqnarray}
    M^1_X &=& \frac{1}{n}\sum_i X_i \hspace{1em} \text{(first moment)}\\
    Var(M^1_X)  &=& Var(\frac{1}{n}\sum_i X_i)\\
    & =& \frac{1}{n^2} Var(\sum_i X_i) = \frac{nVar(X_i)}{n^2} = \frac{\sigma^2}{n}.\nonumber
\end{eqnarray}
\vspace{-1mm}
This is one of the reasons why the circular pattern is more robust to the boundary blur effect than the checkerboard, whose control points are directly obtained from the single point. This finding will be demonstrated via a set of experiments in \cref{sec:Comp.Est.}.

%-------------------------------------------------------------------------% 
\vspace{-1mm}
\subsection{Calibration using unbiased estimator}

The calibration process is divided into two stages. First, we establish the initial value of the intrinsic parameter in closed form using \citet{PAMI-2000-zhang}'s method, assuming zero distortion. The initial value is poor as it does not consider the distortion parameter, requiring refinement through optimization. In the optimization process, we minimize the reprojection error, which is the squared difference between the observed position of the control point and the estimated position of the control point using the unbiased estimator as:
\begin{align}
    \bs{K}, \bs{D} &= \argmin_{\bs{K},\bs{D},{\bs{E}^{1,2,...n}}} \sum_{j=1}^n \sum_{k=1}^m\left\|\bar{\bs{p}}_i^{jk}-\hat{\bs{p}}_i^{jk}\right\|^2,\\
    \hat{\bs{p}}_i^{jk} &= \text{UnbiasedEstimator}(\bs{K}, \bs{E}^j, \bs{Q}^{jk}_w),
\end{align}
where the $\bar{\bs{p}}_i^{jk}$ is the $k^{th}$ control point in the $j^{th}$ image, and the $\hat{\bs{p}}_i^{jk}$ is the estimated control point corresponding to $\bs{p}_i^{jk}$. The circle in target plane corresponding to $\bs{p}_i^{jk}$ is $\bs{Q_w}^{jk}$.
The above optimization problem is solved with the Ceres Solver~\cite{Ceres_Solver}. 

\section{Results}
\label{sec:results}
\begin{figure}[b]
    \centering
    \includegraphics[trim=90 10 90 10,clip,width=0.99\columnwidth]{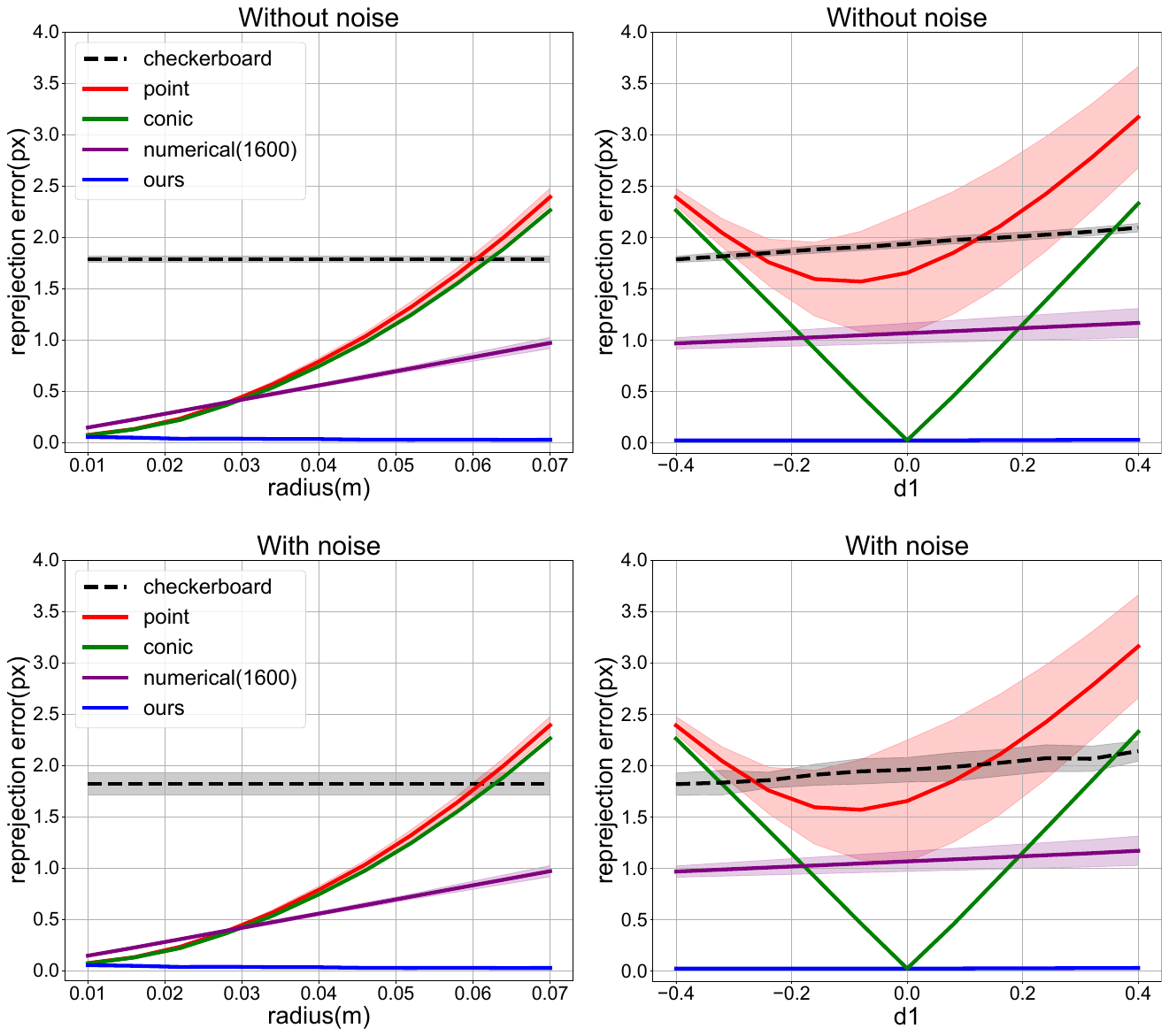}
    \caption{\textbf{Reprojection error comparison in synthetic images.} 
    We examined the consistency and magnitude of the reprojection errors of each method as we varied the radius of the circles and the amount of distortion, represented as $d_1$. In the graphs, each curve represents the mean reprojection error and the envelope represents the corresponding standard deviation boundary. The first row uses raw images and the second row applies the Gaussian blur to the images to check robustness. Unlike other circular pattern methods, our method is unbiased, resulting in near-zero errors regardless of distortion and radius changes. }

    \label{fig:unbiased_merge}
\end{figure}
%--------------------------------------------------------------------------------%

\begin{table*}[t!]
    \centering
    \caption{
    \textbf{Calibration results of synthetic images.} 
    This table shows the mean and standard deviation of the calibration results. We generated 100 synthetic images and obtained intrinsic parameters using each method from 30 randomly selected images. By repeating this action 30 times, we were able to calculate the mean and standard deviation of the calibration results corresponding to each method. Our method shows the closest result to the ground truth value, and the variance is dramatically lower than other methods. Meanwhile, other circular pattern methods, such as point-based and conic-based methods, converge to inappropriate values despite using the same control points as ours.
    More iterations allow the numerical method to behave unbiased while the computational time significantly grows.
    The checkerboard method, which is also unbiased, converges to ground truth values; however, its variance is significant due to measurement noise. Significantly, the measurement accuracy deteriorates further, and almost half of the images are not detected when the image quality is low due to Gaussian blur. }
    \label{tab:cal_results1}
    \resizebox{0.99\textwidth}{!} {
    \begin{tabular}{c|ccccccc|ccccccc}    
        \hline
        &\multicolumn{7}{c|}{Low distortion ($d_1=-0.2$)}&\multicolumn{7}{c}{High distortion ($d_1=-0.4$)}\\
        \hline
        Methods & $f_x$ & $f_y$ &  $c_x$ & $c_y$&  $d_1$ & \#fail &runtime(s) &$f_x$ & $f_y$ &  $c_x$ & $c_y$&  $d_1$ & \#fail &runtime(s) \\
        \hline
        GT & 600 & 600 & 600 & 450 & -0.2& & & 600 & 600 & 600 & 450 & -0.4& & \\
        checkerboard &   599.8$\pm $0.37	&599.8$\pm $0.38	&600.3$\pm $0.21	&449.9$\pm $0.41	&-0.20$\pm $0.001 & 0.0 & 0.3
        & 600.1$\pm $0.22	&600.2$\pm $0.21	&599.9$\pm $0.15	&450.1$\pm $0.12	&-0.40$\pm $ 0.001& 0.0 & 0.4\\
        point based & 598.9$\pm $0.28	&598.9$\pm $0.28	&600.0$\pm $0.29	&450.0$\pm $	0.43&-0.20$\pm $0.001&0.0 &0.3
        &603.1$\pm $0.82	&603.1$\pm $0.79	&599.9$\pm $0.53	&450.3$\pm $0.4	&-0.41$\pm$0.003	&0.0 & 0.4\\
        conic based &603.5$\pm$0.41	&603.5$\pm$0.38	&600.1$\pm$0.24	&449.8$\pm$0.25	&-0.21$\pm$0.002	&0.0 & 0.7
        &606.9$\pm$0.80	&606.9$\pm$0.79	&599.7$\pm$0.74	&450.4$\pm$0.51	&-0.41$\pm$0.005	&0.0 & 0.9\\
        numerical(25) &600.6$\pm$0.28	&600.4$\pm$0.30	&600.0$\pm$0.3	&449.7$\pm$0.22	&-0.20$\pm$0.001	&0.0 & 2.0
        &598.4$\pm$0.71	&597.9$\pm$0.75	&600.9$\pm$0.72	&449.7$\pm$0.36	&-0.40$\pm$0.002	&0.0 & 2.7\\
        numerical(1600) &600.1$\pm$0.07	&600.0$\pm$0.07	&\textbf{600.0}$\pm$\textbf{0.05}	&449.9$\pm$0.06	& \textbf{-0.20}$\pm$\textbf{0.000}	&0.0 & 75.5
        &599.9$\pm$0.1	&599.7$\pm$0.14	&600.2$\pm$0.18	&450.0$\pm$0.06	&\textbf{-0.40}$\pm$\textbf{0.001}	&0.0 & 86.6\\
        ours & \textbf{600.0}$\pm$ \textbf{0.06} &\textbf{600.0}$\pm$\textbf{0.06}  &\textbf{600.0}$\pm$\textbf{0.05}  &\textbf{450.0}$\pm$\textbf{0.05} &\textbf{-0.20}$\pm$\textbf{0.000} &0.0 & 2.0
        &\textbf{599.9}$\pm$\textbf{0.09} &\textbf{599.9}$\pm$\textbf{0.10} &\textbf{600.0}$\pm$\textbf{0.03} &\textbf{450.0}$\pm$\textbf{0.03} &\textbf{-0.40}$\pm$\textbf{0.001} &0.0 & 2.0\\
        \hline
        &\multicolumn{14}{c}{Gaussian blur ($\sigma = 2$)}\\
        \hline
        GT & 600 & 600 & 600 & 450 & -0.2 & & & 600 & 600 & 600 & 450 & -0.4& &\\
        checkerboard &598.7$\pm$1.99   &599.1$\pm$1.84   &599.3$\pm$2.00  &450.5$\pm$1.52  &-0.20$\pm$0.013   &14.2  &0.2
        &596.7$\pm$2.83   &597.1$\pm$2.70   &599.6$\pm$1.98   &451.6$\pm$2.02   &-0.36$\pm$0.026  &15.8 &0.2\\
        point based&598.7$\pm$0.29	&598.7$\pm$0.28	&600.0$\pm$0.32	&449.9$\pm$0.25	&-0.20$\pm$0.003	&0.0  & 0.4
        &603.3$\pm$0.65	&603.3$\pm$0.63	&600.4$\pm$0.64	&450.4$\pm$0.31	&-0.41$\pm$0.002	&0.0 & 0.4 \\
        conic based &603.6$\pm$0.47	&603.6$\pm$0.46	&600.0$\pm$0.28	&449.8$\pm$0.18	&-0.20$\pm$0.002	&0.0 & 0.7
        &607.6$\pm$1.57	&607.7$\pm$1.57	&599.9$\pm$0.60	&450.6$\pm$0.43	&-0.41$\pm$0.004	&0.0& 0.7\\
        numerical(25) & 600.5$\pm$0.26 &600.3$\pm$0.26 &600.1$\pm$0.27 &449.6$\pm$0.18 &\textbf{-0.20}$\pm$\textbf{0.000} &0.0 & 2.0 
        & 598.4$\pm$0.90 &598.0$\pm$0.92 &600.8$\pm$0.65 &449.6$\pm$0.35 &\textbf{-0.40$\pm$0.001} &0.0 & 2.8\\
        numerical(1600) &600.1$\pm$0.09 &600.0$\pm$0.08 &\textbf{600.0}$\pm$\textbf{0.06} &449.9$\pm$0.07 &\textbf{-0.20}$\pm$\textbf{0.000} &0.0 & 75.5
        &599.9$\pm$0.15 &599.7$\pm$0.15 &600.2$\pm$0.18 &450.1$\pm$0.06 &\textbf{-0.40$\pm$0.001} &0.0 & 84.7\\
        ours & \textbf{600.0}$\pm$\textbf{0.07}	&\textbf{600.0}$\pm$\textbf{0.07}	&\textbf{600.0}$\pm$\textbf{0.06}	&\textbf{450.0}$\pm$\textbf{0.05}	&\textbf{-0.20}$\pm$\textbf{0.000}	&0.0 & 2.0
        &\textbf{599.9$\pm$0.07}	&\textbf{599.9$\pm$0.08}	&\textbf{600.0$\pm$0.04}	&\textbf{450.0$\pm$0.03}	&\textbf{-0.40$\pm$0.001}	&0.0 &2.0\\
        
        \hline
    \end{tabular}}
    
    % \vspace{-5mm}
\end{table*}
\subsection{Comparison of estimators}
\label{sec:Comp.Est.}

We compared our unbiased estimator with the checkerboard method and other existing estimators, such as point-based, conic-based, and numerical method~\cite{PAMI-2006-kannala}, as defined below:

\begin{align}
    \text{Point-based} &: \hat{\Tilde{\bs{p}}}_i = \bs{K}D(\bs{E}\bs{Q}_w^{-1}[0,0,1]^{\top}),\\
    \text{Conic-based} &: \hat{\Tilde{\bs{p}}}_i = \bs{K}D(\bs{E}\bs{Q}_w^{-1}\bs{E}^T[0,0,1]^{\top}),\\
    \text{Numerical} &: \hat{\Tilde{\bs{p}}}_i =\text{Eq. (19) in \cite{PAMI-2006-kannala}}\\
    \text{Ours} &: \text{\cref{alg:unbiased}} \nonumber
\end{align}
Numerical($n$) means the iteration step is $n$.

Most of the existing calibration algorithms use the point-based estimator, which disregards the conic geometry and directly matches the center of the shape in the distorted image to the center of the circle in the target. The conic-based estimator compensates for the bias introduced by the perspective transformation. However, the conic-based estimator still obtains the distortion bias. The numerical method, which is theoretically unbiased, also shows considerable error induced by numerical integration.
These limitations are well described in \cref{fig:unbiased_merge}. Each data point indicates the average of reprojection errors in fixed 24 scenes, and we illustrated the error graph with standard deviations.

Our moment-based unbiased estimator maintains very small reprojection errors regardless of radius size and distortion. However, the errors of other methods are significant, while the errors of circular pattern methods such as conic and point gradually increase as the radius or distortion increases. Although the amount of distortion does not greatly influence the result of the checkerboard method, it has the inherent error caused by the inaccurate control point detector compared to the circular pattern. The control point of the checkerboard is the corner of the squares, and due to the discontinuity of the pixel, the corner position should be approximated by interpolation near the pattern boundary. In contrast, the control point of the circular pattern is obtained from the average of thousands of inner points of the circle, resulting in the precise position in decimal places.

Because the conic and point-based estimators have biases resulting from the conic geometry, the error increases as the circle size and distortion increase. The two estimators have slightly different tendencies. The conic-based estimator has a bias due to the distortion, not the perspective transformation. Hence, the reprojection error depends only on the absolute value of the distortion coefficient. In comparison, the point based estimator has both perspective and distortion biases. The perspective bias causes a high standard deviation and asymmetry in the error plot.

%--------------------------------------------------------------------------------%
\subsection{Calibration accuracy of synthetic images}
\label{sec:exp_syn}

We conducted experiments on synthetic images to evaluate how our unbiased estimator improves calibration performance. We set the \ac{GT} values for $f_x$, $f_y$, $c_x$, and $c_y$ to make the \ac{FOV} approximately 90 degrees. We prepared two scenarios for the distortion coefficients from \eqref{eq:def_dist}: one with low distortion ($d_1$ = -0.2) and one with high distortion ($d_1$ = -0.4). In the latter case, the distortion function is not invertible within the given \ac{FOV} range, so we slightly adjusted $d_2$ to make it invertible, and we only evaluated $d_1$ in our analysis since $d_2$ is as small as negligible. We prepared 100 images and performed calibration on a set of 30 randomly selected images. We repeated this process 30 times to calculate the averages and standard deviations shown in \cref{tab:cal_results1}.

As a result, our method consistently produces the best results in all cases. Our approach benefits from an accurate mean and is particularly valuable for its low variance.
Since the estimator of the checkerboard pattern is unbiased, it converges closely to the ground truth in the absence of noise. However, the calibration results show significant variance from the imprecise control point measurements, as shown in the experiment in \cref{sec:Comp.Est.}. This tendency increases when noise is introduced into the images, and control point detection fails in many cases, resulting in worse performance.

In the case of point-based and conic-based methods using a circular pattern, the measurements are accurate and robust. However, due to the biased estimator, they perform worse than the checkerboard pattern. As the distortion increases, the bias becomes more pronounced, leading to poor calibration results. The numerical method has a trade-off between accuracy and runtime, and it requires significant time to achieve meaningful results. 
Our method, which addresses these limitations, exhibits higher accuracy, efficiency, and robustness than the checkerboard and the others. Note that one of the critical contributions of this paper is to prove the value of circular patterns, which have more informative features but have been underutilized due to algorithmic limitations.

%--------------------------------------------------------------------------------%
\subsection{Calibration accuracy of real images}

\begin{figure}[t!]
    \centering
    \includegraphics[trim=130 50 110 50, clip,width=1.0\columnwidth]{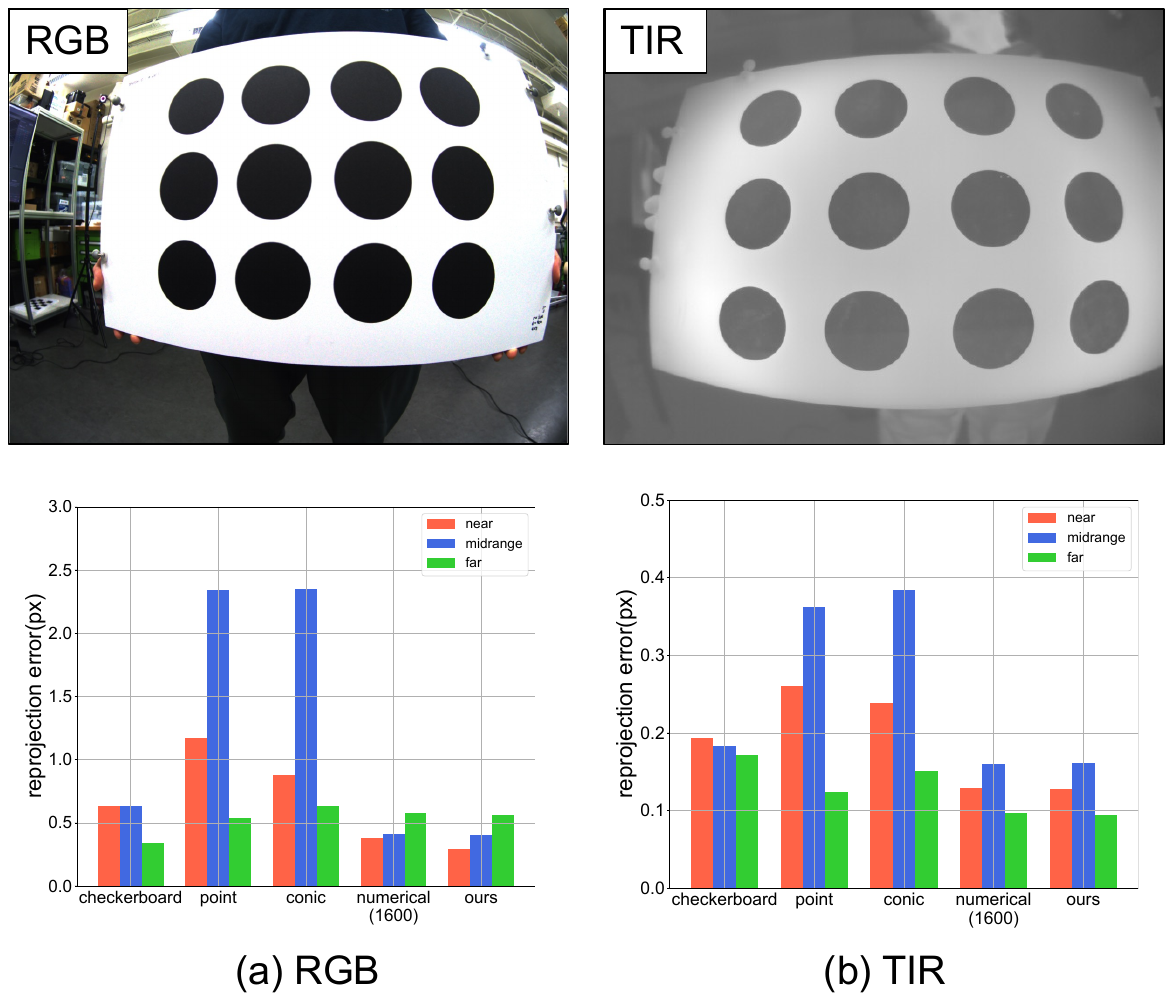}
    \vspace{-5mm}
    \caption{\textbf{Real world experiment: Reprojection error.} 
 The first row shows sample images from (a) RGB and (b) TIR cameras. The second row shows the distribution of the reprojection error divided by the distance from the camera to the target. The error distribution of the biased methods, such as point and conic-based methods, varies greatly with distance. The unbiased methods, such as checkerboard and ours, show a low and uniform error distribution. In TIR images, our method significantly outperforms the checkerboard method due to the robust control point of circular patterns.
     }
    \label{fig:dist_rpe}
    \vspace{-5mm}
\end{figure}

We further evaluated our method using real images captured from both RGB and \ac{TIR} cameras. Another evaluation metric is required to extend the experiment to the real world since there is no way to find the \ac{GT} intrinsic parameter of real cameras. Instead of comparing the intrinsic parameter directly, we evaluated the distribution of reprojection errors and the relative translation and rotation value between each target in different scenes. To highlight clear differences among the methods, we utilized two types of cameras with high distortion. One is an RGB camera, Trition 5.4 MP, whose resolution is 1200$\times$930. The other is a \ac{TIR} camera, FLIR A65, whose resolution is 640$\times$512. The sample images of these two cameras are shown in \figref{fig:dist_rpe}. Since the \ac{TIR} is limited in recognizing colored patterns and only distinguishes objects by infrared energy we utilized a printed circuit board (PCB) composed of squares or circles with different heat conductivity introduced in \cite{RAL-2019-shin}.

\noindent \textbf{Distribution of reprojection errors.} We expected calibration results to be better when the total reprojection error is small and the error is uniformly distributed. We collected 24 images from three different distances (i.e., near, midrange, and far) to investigate these two aspects. As summarized in \cref{fig:dist_rpe}, point-based and conic-based methods result in high reprojection errors and significant differences in error values between different distances, which implies that the calculated intrinsic parameters do not explain the projection model consistently across all distances. On the contrary, the checkerboard method and our method, whose unbiased estimators show low reprojection error at all distances. The checkerboard and our methods show similar performance in the RGB camera. Meanwhile, our methods outperform the checkerboard method in the TIR camera due to accurate control point detection. As shown in \secref{sec:exp_syn} the accuracy of the numerical method converge to our method while the number of iteration step is increasing. Numerical(1600) shows similar results to ours, while it takes almost five minutes to converge. More information such as the distribution of the error vectors on the 2D image is described in \bl{\apref{sec:rep_vector}}.

\begin{figure}[b]
    \centering
    \includegraphics[trim=110 170 130 110,clip,width=1.0\columnwidth]{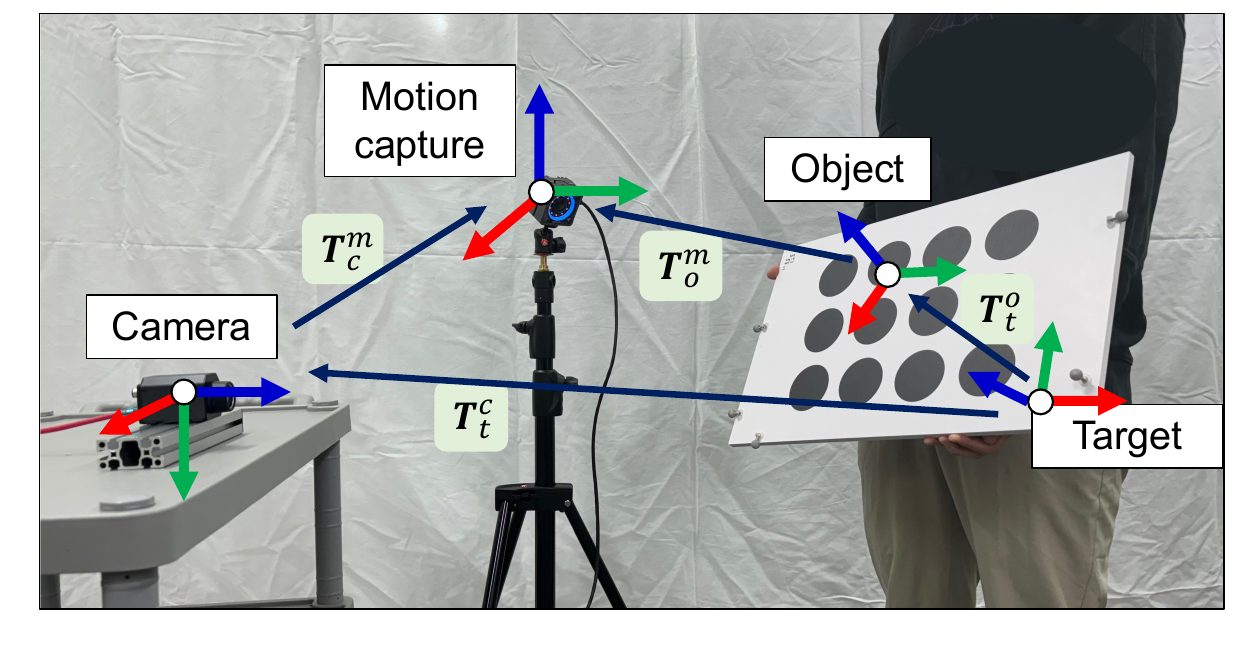}
    \vspace{-10mm}
    \caption{\textbf{Experiment setup.} We utilize the motion capture system to develop the ground truth 6D pose of the target. The $\bs{T}_c^m \bs{T}^c_t = \bs{T}^m_o \bs{T}^o_t$ relationship is satisfied.}
    \label{fig:exp_geo}
\end{figure}

\begin{table}[t]
    \centering
    \caption{\textbf{Real world experiment: 6D pose error.} The rotation and translation error between the target pose obtained from the calibration results and the motion capture system. Our method outperforms other methods in the rotation part. The checkerboard method shows comparable results in the translation part on RGB cameras; however, it malfunctions on the TIR camera.}
    \label{tab:pose_error}
    \resizebox{0.99\columnwidth}{!} {
    \begin{tabular}{c|c|c|c|c}
        \hline
        & \multicolumn{2}{c|}{RGB} & \multicolumn{2}{c}{TIR}\\
        \hline
        Methods & Rotation & Translation & Rotation & Translation \\
        \hline
        \hline
        checkerboard& 0.32\degree & \textbf{2.5} \mm & 1.63\degree & 16.5 \mm \\
        point-based& 0.36\degree & 4.2 \mm & 0.50\degree & 3.8 \mm \\
        conic-based& 0.31\degree & 2.7 \mm & 0.47\degree & 3.4 \mm \\
        numerical(1600)& 0.26\degree & 3.8 \mm & \textbf{0.45}\degree & 3.3 \mm \\
        ours& \textbf{0.24}\degree & \textbf{2.5} \mm & \textbf{0.45}\degree & \textbf{3.1} \mm \\
        \hline
    \end{tabular}}
\end{table}

\noindent \textbf{6D pose errors.} To validate the un-projection performance of each method, we investigated the relative 6-\ac{DoF} pose error. Each camera captured $20$ different scenes and we conducted the calibration with these images. During the calibration, the pose of the target in each scene is also optimized, and these poses are used to evaluate the calibration accuracy. The \ac{GT} value of the relative pose is obtained by the motion capture system by OptiTrack. Let $ \bs{T}_{t_i}^c$ be a $SE(3)$ matrix from the $i^{\text{th}}$ target coordinates to the camera coordinate, and  $ \bs{T}_{o_i}^m$ be a $SE(3)$ matrix from the $i^{\text{th}}$ object coordinates to the motion capture system coordinate. Note that
the object frame is the frame attached to the target plane. However, the two frames are not identical since the motion capture system defines another coordinate system for the target as \cref{fig:exp_geo}.

It is impossible to evaluate the relative pose directly without the coordinate transformation matrices $\bs{T}^o_t$ and $\bs{T}^m_c$. Therefore, we first calculated the transformation matrices, and the final error is defined as 
\begin{equation}
    \label{eq:motion6}
    error = \sum_{i=1}^n||\bs{T}^m_{o_i} \ominus \hat{\bs{T}^m_c} \bs{T}^c_{t_i} (\hat{\bs{T}^o_t})^{-1}||
\end{equation}
with the estimated value $\hat{\bs{T}^m_c}$ and $\hat{\bs{T}^o_t}$. The detail of estimating $\bs{T}^o_t$ and $\bs{T}^m_c$ is described in \bl{\apref{Appendix:ax=xb}}
According to \tabref{tab:pose_error}, our method outperforms the entire case, and the checkerboard method shows lower performance at the un-projection task despite the low reprojection error compared to the point and conic-based methods.

\section{Conclusion}
\label{sec:conclusion}
In this paper, we introduce the closed-form $n^{\text{th}}$ moment of conics and prove that the centroid of a distorted ellipse is expressed by a linear combination of the original moments. Based on these results, we developed an unbiased estimator of circular patterns under distortion for camera calibration. Using this estimator, circular pattern-based calibration overcomes its algorithmic limitation and outperforms checkerboard-based calibration.

\vspace{5mm}
\noindent \textbf{Acknowledgment} This work was supported by the NRF grant funded by MSIP (2020R1C1C1006620) and IITP grant funded by MSIT (No.2022-0-00480).
{
    \small
    \bibliographystyle{ieeenat_fullname}
    \bibliography{string-long,ref}
}

\clearpage
\setcounter{page}{1}
\onecolumn
\begin{center}
    \Large
    \textbf{Unbiased Estimator for Distorted Conics in Camera Calibration}
        
    \vspace{0.4cm}
    \large
    Supplementary Material
\end{center}

\section{Proof of Theorem and lemma in \cref{sec:method}}

\subsection{Proof of Theorem 1}
\label{Appendix:Theorem1}

\begin{proof}
\begin{eqnarray}
  [x',y'] &=& D(x,y) = [f(x,y), g(x,y)]\\
  \bs{J} &=& Jacob(D) = \begin{bmatrix}
        \frac{\partial f}{\partial x} && \frac{\partial f}{\partial y}\\
        \frac{\partial g}{\partial x} && \frac{\partial g}{\partial y}
        \end{bmatrix},\\
    dA_{x'y'} &=&    \sqrt{det(\bs{G})}dA_{x'y'} =\sqrt{det(\bs{J}^{\top}\bs{J})}dA_{x'y'}=det(\bs{J})dA_{x'y'},\\
  |A_{x'y'}|M_{x'y'}^{m,n} &=& \int (x')^m(y')^n dA_{x'y'} ,\\
    &=& \int f(x,y)^n g(x,y)^m \det(\bs{J})dA_{xy}. 
\end{eqnarray}
The positive definite matrix $\bs{G}$ is called The First Fundamental Form or Riemannian Metric. Since $f(x,y)$, $g(x,y)$, and $det(\bs{J})$ are polynomial functions of $x$ and $y$, $M_{x'y'}^{n,m}$ can be expressed by a linear combination of $M_{xy}^{i,j}$.
\begin{eqnarray}
  |A_{x'y'}|M_f^{n,m}  &=& \int \sum_{ij}c_{ij} x^iy^jdA_{xy}\\ &=& |A_{xy}|\left[ \frac{1}{|A_{xy}|} \int \sum_{ij}c_{ij} x^iy^jdA_{xy} \right]\\
  &=&  |A_{xy}| \sum_{ij}c_{ij} \left[ \frac{1}{|A_{xy}|} \int x^iy^jdA_{xy} \right] \\
  &=& |A_{xy}|\sum_{i=0}^{p}\sum_{j=0}^{q} c_{ij}M_{xy}^{i,j}.
\end{eqnarray}
\end{proof}
\subsection{Proof of Theorem 2}
\label{Appendix:Theorem2}
\begin{proof}
  \begin{eqnarray}
    x_n &=& \cos\alpha x_s -\sin\alpha y_s \\
    y_n &=& \sin\alpha x_s + \cos\alpha y_s \\
    s_n &=& x_n^2+y_n^2 = x_s^2 + y_s^2 = s_s \\ 
    dA_n &=& \det(\bs{J})dA_s\nonumber\\ &=& (\cos^2\alpha+\sin^2\alpha)dA_s = dA_s\\
    \therefore |A_n| &=& |A_s|\\
    \bs{v}_n^r 
    &=&\begin{bmatrix}
            \frac{1}{|A_n|} \int  x_ns_n^rd A_n \\ \frac{1}{|A_n|} \int y_ns_n^rd A_n \\\frac{1}{|A_n|} \int s_n^rd A_n 
    \end{bmatrix} = \begin{bmatrix}
        \frac{1}{|A_s|} \int (\cos\alpha x_s -\sin\alpha y_s) s_s^rd A_s\\ \frac{1}{|A_s|} \int (\sin\alpha x_s + \cos\alpha y_s) s_s^rd A_s \\\frac{1}{|A_s|} \int s_s^rd A_s 
        \end{bmatrix}\\
    &=& \begin{bmatrix}
            \cos\alpha & -\sin\alpha &0\\
            \sin\alpha & \cos\alpha & 0\\
            0 & 0 & 1\\
    \end{bmatrix} 
    \begin{bmatrix}
        \frac{1}{|A_s|} \int  x_ss_s^rd A_s \\ \frac{1}{|A_s|} \int y_ss_s^rd A_s \\\frac{1}{|A_s|} \int s_s^rd A_s 
    \end{bmatrix} 
  \end{eqnarray}  
\end{proof}
\subsection{Proof of Lemma 3}
\label{Appendix:Lemma3}
\begin{proof}
\begin{eqnarray}
    I^{m,n} 
    &=& \frac{1}{2\pi}\int_0^{2\pi} \cos^{m-1}\theta\cos\theta\sin^n\theta d\theta\\
    & =&\frac{1}{2\pi} \left[\frac{1}{n+1}\cos^{m-1}\theta\sin^{n+1}\theta\right]_0^{2\pi}+ \frac{m-1}{n+1}\frac{1}{2\pi}\int \cos^{m-2}\theta \sin^{n+1}\theta \sin\theta d\theta \\
    \label{eq:3-1}
    &=& \frac{m-1}{n+1} I^{m-2, n+2}\\
    Similarly, \nonumber \\
    \label{eq:3-2}
    I^{m,n} &=& \frac{n-1}{m+1} I^{m+2, n-2}
\end{eqnarray}
By Eqs. \eqref{eq:3-1} and \eqref{eq:3-2}, we can reduce m or n to zero or one. However, $I^{1,n}$ and $I^{m,1}$ is always zero as follows. 
\begin{eqnarray}
    I^{1,n} &=& \frac{1}{2\pi}\int_0^{2\pi} \cos\theta\sin^n\theta d\theta= \left[\frac{1}{n+1}\sin^{n+1}\theta\right]_0^{2\pi}=0,\\
    I^{m,1} &=& \frac{1}{2\pi}\int_0^{2\pi} \cos^m\theta\sin\theta d\theta= \left[-\frac{1}{m+1}\cos^{m+1}\theta\right]_0^{2\pi}=0.
\end{eqnarray}
Therefore, we only need to consider the even case when $m$ is $2i$ and $n$ is $2j$. Otherwise, $I^{m,n}=0$.

\noindent$I^{0,n}$ also has a reduction formula as
\begin{eqnarray}
    I^{0,n} 
    &=& \frac{1}{2\pi}\int_0^{2\pi} \sin^n\theta d\theta =\frac{1}{2\pi} \int_0^{2\pi} \sin^{n-1}\theta \sin\theta d\theta\\
    &=& \frac{1}{2\pi}\left[-\cos\theta\sin^{n-1}\theta\right]_0^{2\pi} + \frac{n-1}{2\pi}\int_0^{2\pi} \sin^{n-2}\theta \cos^2\theta d\theta\\
    & =&  \frac{n-1}{2\pi}\int_0^{2\pi} \sin^{n-2}\theta (1-\sin^2\theta) d\theta\\
    &=& -(n-1)I^{0,n} + (n-1)I^{0,n-2}, \\
    \label{eq:3-3}
    \therefore I^{0,n} &=& \frac{n-1}{n}I^{0,n-2}.
\end{eqnarray}
Using Eqs. \eqref{eq:3-1} and \eqref{eq:3-3}, the analytic solution of $I^{2i, 2j}$ is obtained.
\begin{eqnarray}
I^{2i,2j} &=& \frac{2i-1}{2j+1}\frac{2i-3}{2j+3}\cdots\frac{1}{2j+2i-1}I^{0,2j+2i}\\
    & =& \frac{2i-1}{2j+1}\frac{2i-3}{2j+3}\cdots\frac{1}{2j+2i-1}\frac{2i+2j-1}{2i+2j}\frac{2i+2j-3}{2i+2j-2}\cdots\frac{1}{2}\\
    &=& \frac{(2i-1)(2i-3)\cdots(1)(2j-1)\cdots(1)}{(2i+2j)(2i+2j-2)\cdots(2)}\\
     &=& \frac{(2i-1)(2i-3)\cdots(1)(2j-1)\cdots(1)}{(2i+2j)(2i+2j-2)\cdots(2)}\frac{(2i)(2i-2)\cdots(2)(2j)(2j-2)\cdots(2)}{(2i)(2i-2)\cdots(2)(2j)(2j-2)\cdots(2)}\\
     \label{eq:AIij-1}
    & =& \frac{(2i)!(2j)!}{(i+j)!i!j!} \frac{1}{2^{2i+2j}}
\end{eqnarray}
The \equref{eq:AIij-1} is a symmetric equation of $i$ and $j$. The factorial term such as $(i+j)!$ readily induces numerically unstable; therefore, we convert the factorial term to a combination term such as $\binom{i+j}{i}$. 
\begin{equation}
    \frac{(2i)!(2j)!}{(i+j)!i!j!} \frac{1}{2^{2i+2j}}
    = \frac{(2i)!(2j)!}{(2i+2j)!}\frac{(2i+2j)!}{(i+j)!(i+j)!}\frac{(i+j)!}{i!j!} \frac{1}{2^{2i+2j}}= \frac{\binom{2i+2j}{i+j}\binom{i+j}{i}}{\binom{2i+2j}{2i}2^{2i+2j}}.
\end{equation}
Another advantage of combination terms is that we can develop a combination matrix in advance using Pascal's triangle.

\end{proof}
\subsection{Proof of Lemma 4}
\label{Appendix:Lemma4}
\begin{proof}
    \begin{eqnarray}
        x_0 &=& ar\cos\theta,\\
        y_0 &=& br\sin\theta,\\
        M^{m,n}_0 &=& \frac{1}{\pi ab}\int x_0^m y_0^n d A_0\\
        &=& \int_0^{2\pi}\int_0^1 a^mb^n r^{m+n}\cos^m\theta\sin^n\theta \ rab \ drd \theta\\
        & =& \frac{a^{m+1}b^{n+1}}{\pi ab(m+n+2)} \int_0^{2\pi} \cos^m\theta\sin^n\theta d\theta\\
        & =& \frac{a^mb^n}{1+(m+n)/2} I^{m,n}.
    \end{eqnarray}
\end{proof}
\subsection{Proof of Theorem 5}
\label{Appendix:Theorem5}
\begin{proof}
    \begin{eqnarray}
        dA_s &=& dA_0,\\
        \frac{1}{|A_s|}\int (x_s^2+y_s^2)^r dA_s  
        &=& \frac{1}{|A_s|} \sum_{k=0}^{r}\binom{r}{k}\int x_s^{2k}y_s^{2r-2k} dA_0      \\
        &=& \frac{1}{|A_s|} \sum_{k=0}^{r}\binom{r}{k}\int(x_0+t_x)^{2k}(y_0+t_y)^{2r-2k} dA_0\\
        &=& \sum_{k=0}^{r} \binom{r}{k} \sum_{p=0}^{2k}\sum_{q=0}^{2r-2k}\binom{2k}{p} \binom{2r-2k}{q}t_x^{2k-p}t_y^{2r-2k-q}M_0^{pq}\\
        &=& \sum_{p=0}^{2r}\sum_{q=0}^{2r-p} M_0^{pq} \sum_{k=\lceil p/2 \rceil}^{\lfloor r-q/2 \rfloor} \binom{r}{k} \binom{2k}{p}\binom{2r-2k}{q}t_x^{2k-p}t_y^{2r-2k-q}    
    \end{eqnarray}
    Since $M_0^{pq}$ is zero when p or q is odd number, we can rewrite the above equation using $p=2i$ and $q=2j$. Using the above equations, we obtain Eqs. \eqref{eq:t5-1}--\eqref{eq:t5-3}.
    \begin{eqnarray}
       % \bs{v}_s^r &\triangleq& \begin{bmatrix} v_0, v_1, v_2 \end{bmatrix}^\top \nonumber\\
       \label{eq:t5-1}
        \bs{v}_s^r[2] = \frac{1}{|A_s|}\int (x_s^2+y_s^2)^r dA_s
        &=& \sum_{i=0}^{r}\sum_{j=0}^{r-i}M_0^{2i,2j} \sum_{k=i}^{r-j} \binom{r}{k} \binom{2k}{2i}\binom{2r-2k}{2j}t_x^{2k-2i}t_y^{2r-2k-2j}  \\
        \label{eq:t5-2}
        \bs{v}_s^r[0] = \frac{1}{|A_s|}\int x_s(x_s^2+y_s^2)^r dA_s
        &=& \frac{1}{|A_s|} \sum_{k=0}^{k=r}\binom{r}{k}\int(x_0+t_x)^{2k+1}(y_0+t_y)^{2r-2k} dA_0\nonumber \\
        &=& \sum_{k=0}^{r} \binom{r}{k} \sum_{p=0}^{2k+1}\sum_{q=0}^{2r-2k}\binom{2k+1}{p} \binom{2r-2k}{q}t_x^{2k-p+1}t_y^{2r-2k-q}M_0^{pq}\nonumber \\
        &=& \sum_{i=0}^{r}\sum_{j=0}^{r-i} M_0^{2i,2j} \sum_{k=i}^{r-j} \binom{r}{k} \binom{2k+1}{2i}\binom{2r-2k}{2j}t_x^{2k-2i+1}t_y^{2r-2k-2j} \\
        Similarly,\hspace{3cm}&& \nonumber\\
        \label{eq:t5-3}
        \bs{v}_s^r[1] = \frac{1}{|A_s|}\int y_s(x_s^2+y_s^2)^r dA_s
        &=& \sum_{i=0}^{r}\sum_{j=0}^{r-i} M_0^{2i,2j} \sum_{k=i}^{r-j} \binom{r}{k} \binom{2k}{2i}\binom{2r-2k+1}{2j}t_x^{2k-2i}t_y^{2r-2k-2j+1} 
    \end{eqnarray}
\end{proof}

\section{Derivation details}

\subsection{(\cref{sec:momentum}) Geometric Feature of Ellipse}
\label{Appendix:Ellipse_feature}
When the $\bs{Q}$ represents an ellipse, the geometric features of the ellipse (i.e., center point and major/minor axis length) could be obtained from the matrix $\bs{Q}$ as follows.
\begin{eqnarray}
  \bs{Q}&\triangleq& \begin{bmatrix}
      a&b&d\\
      b&c&e\\
      d&e&f
  \end{bmatrix}\\
  h_1 &=& (ac-b^2), \label{eq:conic1}\\
  h_2 &=& \sqrt{(a-c)^2+4b^2}, \label{eq:conic2}\\
  t_x &=& (be-cd)/h_1, \label{eq:conic3}\\
  t_y & =& (bd-ae)/h_1, \label{eq:conic4}\\
  m_0 & =& \sqrt{\frac{2\det(\bs{Q})}{h_1(a+c-h_2)}}, \label{eq:conic5}\\
  m_1 & = &\sqrt{\frac{2\det(\bs{Q})}{h_1(a+c+h_2)}}, \label{eq:conic6}\\
  \alpha & =& \tan^{-1}\left(\frac{c-a+h_1}{-2b}\right) \label{eq:conic7}.
\end{eqnarray}
Here, $t_x$ and $t_y$ are the center point of the ellipse, and $m_0$ and $m_1$ are the lengths of ellipse axes. The angle between the x-axis and the ellipse axis, whose length is $m_o$, is denoted as $\alpha$ in \figref{fig:moment_geo}

\subsection{(\cref{sec:tracking}) Tracking moment under distortion}
\label{Appendix:Moment_tracking}

The $n_d$ is the number of distortion parameters defined in \equref{eq:def_dist}.  
\begin{eqnarray}
    s &=& x_n^2+y_n^2,\\
    k & =& 1 + d_1s+d_2s^2+d_3s^3+...+d_ns^n = \sum_{i=0}^{n_d}d_is^i \hspace{1cm} (d_0=1),\\
    x_d &=& kx_n\\
    y_d &=& ky_n,\\
    dA_d &=& \sqrt{det(\bs{G})}dA_n\\
    \sqrt{det(\bs{G})} & =& \sqrt{det(\bs{J}^{\top}\bs{J})}=det(\bs{J}) =\begin{vmatrix}
        \frac{\partial x_d}{\partial x_n} & \frac{\partial x_d}{\partial y_n} \\
        \frac{\partial y_d}{\partial x_n} & \frac{\partial y_d}{\partial y_n}
    \end{vmatrix}\\
    &=&\begin{vmatrix}
        k+2x_n^2\frac{\partial k}{\partial s} & 2x_ny_n\frac{\partial k}{\partial s}\vspace{0.2cm},\\
        2x_ny_n\frac{\partial k}{\partial s} & k+2y_n^2\frac{\partial k}{\partial s}
    \end{vmatrix} = k(k+2s\frac{\partial k }{\partial s}) = \sum_{i=0}^{n_d}d_is^i\sum_{i=0}^{n_d}(2i+1)d_is^i.
\end{eqnarray}
Using the above equations, we obtain
\begin{eqnarray}
    \label{eq:Am00}
    \frac{|A_d|}{|A_n|} &=& \frac{1}{|A_n|}\int dA_d = \frac{1}{|A_n|}\int \sqrt{det(\bs{G})}~dA_n = \frac{1}{|A_n|} \int \left(\sum_{i=0}^{n_d}d_is^i\right)\left(\sum_{i=0}^{n_d} (2i+1)d_is^i\right) ~dA_n.\\
    |A_d|M_d^{1,0} &=& \int x_d~ dA_d = \int kx_n\sqrt{det(\bs{G})}~dA_n =\int x_nk^2(k+2s\frac{\partial k }{\partial s})~dA_n\\
    \label{eq:Am10}
    M_d^{1,0} &=& \frac{|A_n|}{|A_d|} \frac{1}{|A_n|}\int x_n\left(\sum_{i=0}^{n_d}d_is^i\right)^2\left(\sum_{i=0}^{n_d} (2i+1)d_is^i\right) ~dA_n.\\
    |A_d|M_d^{0,1} &=& \int y_d~ dA_d = \int y_nk^2(k+2s\frac{\partial k }{\partial s})~dA_n,\\
    \label{eq:Am01}
    M_d^{0,1} &=& \frac{|A_n|}{|A_d|} \frac{1}{|A_n|} \int y_n\left(\sum_{i=0}^{n_d}d_is^i\right)^2\left(\sum_{i=0}^{n_d} (2i+1)d_is^i\right) ~dA_n.
\end{eqnarray}

We can reduce the computational cost as follows.
\small
\begin{eqnarray}
   \left( \sum_{i=0}^{n_d} d_is^i \right) \left( \sum_{i=0}^{n_d} (2i+1)d_is^i \right)  &=& \sum_{i=0}^{n_d}\sum_{j =0}^{n_d} (2i+1)d_id_js^{i+j}\\
    &=& \sum_{r=0}^{2n_d} w_{0r}\cdot s^r, \hspace{0.5cm}\left(w_{0r} = \sum_{i=max(0, r-n_d)}^{min(r,n_d)} (2i+1)d_id_{r-i}\right)\\
    \left( \sum_{i=0}^{n_d} d_is^i \right)^2 \left( \sum_{i=0}^{n_d} (2i+1)d_is^i \right)  &=& \sum_{i, j, k =0}^{n_d} (2i+1)d_id_jd_ks^{i+j+k}\\
    &=& \sum_{r=0}^{3n_d} w_{1r}\cdot s^r \hspace{0.5cm}\left(w_{1r} = \sum_{i=max(0, r-2n_d)}^{min(r,n_d)} (2i+1)d_i\sum_{j=max(0,r-i-n_d)}^{min(r-i,n_d)} d_jd_{r-i-j}\right)
\end{eqnarray}
\normalsize

Then, Eqs. \eqref{eq:Am00} to \eqref{eq:Am01} are rewritten as
\begin{align}
    \frac{|A_d|}{|A_n|} &= \sum_{r=0}^{2n_d} w_{0r} \left[ \frac{1}{|A_n|} \int s^r dA_n\right]\\
    M_d^{1,0} &= \frac{|A_n|}{|A_d|} \sum_{r=0}^{3n_d} w_{1r} \left[ \frac{1}{|A_n|} \int x_ns^r dA_n\right]\\
    M_d^{0,1} &= \frac{|A_n|}{|A_d|} \sum_{r=0}^{3n_d} w_{1r} \left[ \frac{1}{|A_n|} \int y_ns^r dA_n\right]
\end{align}

\subsection{(\cref{sec:tracking}) The centroid of the distorted ellipse on the image plane}
\label{Appendix:ImageCenterPoint}
% In the \cref{sec:preliminary}, we do not consider the skew parameter $\eta$ in intrinsic matrix $\bs{K}$ for convenience since it is negligible in most cameras. In fact, our method is not limited to this assumption and we provide the general derivation in this proof.
% \vspace{2cm}
\begin{eqnarray}
    \begin{bmatrix}
        x_i\\
        y_i\\
    \end{bmatrix}
    &=& \begin{bmatrix}
        f_x& \eta& c_x\\
        0 & f_y & c_y
    \end{bmatrix}\begin{bmatrix}
        x_d\\ y_d \\ 1
    \end{bmatrix}\\
    det(\bs{J}) & =&\begin{vmatrix}
        f_x & \eta\\
        0 & f_y
    \end{vmatrix} = f_xf_y,\\
    |A_i| &=& f_xf_y|A_d|,\\
    \Bar{x_i} \triangleq M_i^{1,0}&=& \frac{1}{|A_i|} \int x_i dA_i = \frac{|A_d|}{|A_i|}\frac{1}{|A_d|}\int (f_x x_d+\eta y_d+ c_x)f_xf_ydA_d \\
    &=& f_xM_d^{1,0}+\eta M_d^{0,1} +c_x\hspace{0.2cm}(\because |A_i| = f_xf_y|A_d|),\\
    \Bar{y_i} \triangleq M_i^{0,1}  &=& f_yM_d^{0,1}+c_y,\\
     \begin{bmatrix}
        \Bar{x_i}\\
        \Bar{y_i}
    \end{bmatrix} &=& \begin{bmatrix}
        f_x & \eta & c_x\\
        0 & f_y & c_y
    \end{bmatrix}\begin{bmatrix}
       \Bar{x_d} \\ \Bar{y_d}\\1
    \end{bmatrix}.
\end{eqnarray}

\section{Experiments details}
\subsection{Characteristics of TIR camera}
\label{Appendix:TIR_Char}
\begin{figure}[b!]
    \centering
    \includegraphics[trim=120 170 120 160, clip,width=0.8\columnwidth]{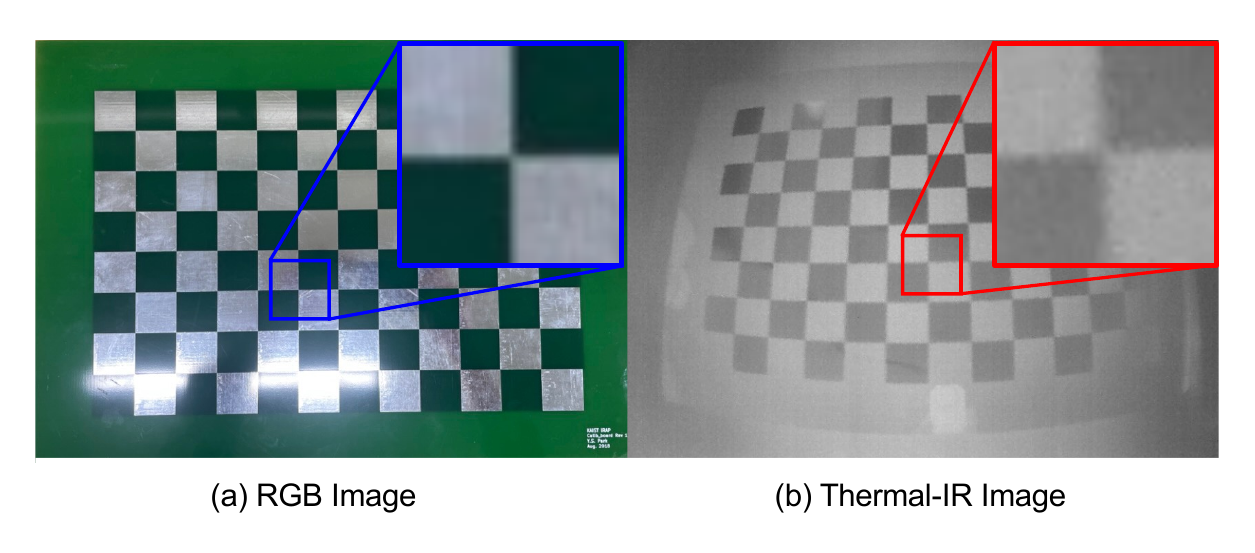}
    \caption{\textbf{The checkerboard pattern captured from the RGB camera (left) and TIR
camera (right)}. Compared to RGB images, the boundary is highly
blunt when captured from TIR. Severe distortion can also be
observed in the thermal images}
    \label{fig:boundaryblur}
\end{figure}
The thermal infrared (TIR) camera is a distinctive vision sensor for extreme environments.
The TIR camera is limited in recognizing colored patterns because it
distinguishes objects by infrared energy, not by color; hence,
a TIR camera needs a particular target for calibration. Some thermal-specific
calibration targets in the literature include a printed circuit board (PCB) composed of different heat conductivity squares \cite{RAL-2019-shin}. However, even with this target, achieving high calibration accuracy is onerous. The thermal images often include low resolution, high distortion, and blunt boundaries, possibly leading to inaccurate control point detection. For example, the temperature discrepancy between two adjacent objects is decreased by conduction and radiation. This phenomenon causes blunt
boundaries as illustrated in \cref{fig:boundaryblur}.

\subsection{Vector representation of reprojection errors on the RGB image }
\label{sec:rep_vector}
\begin{figure}[t]
    \centering
    \includegraphics[trim=100 250 95 180, clip,width=1.0\columnwidth]{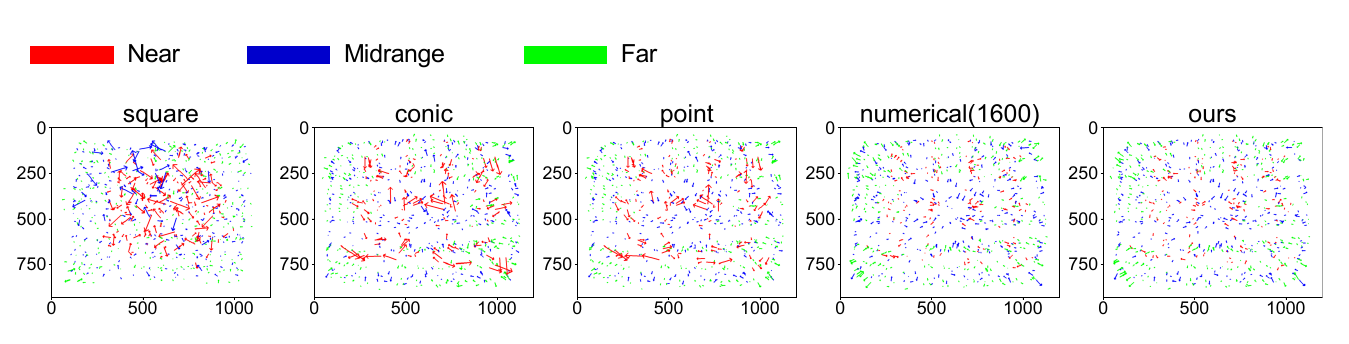}
    \caption{\textbf{Visualization of reprojection error vector.} We collected 60 images and scattered 720 reprojection error vectors of the control points on the image plane. While our method shows low reprojection error across the entire image area, other methods include larger error vectors. For visualization, the error vectors are scaled up 50 times.}
    \label{fig:rep_vec}
\end{figure}
We visualized the reprojection error per the actual distance from the camera to the calibration target in \cref{fig:dist_rpe}. To provide more intuition about the distribution in spatial aspect, we performed the calibration using 20 images and scattered the reprojection error vector on the image plane. \cref{fig:rep_vec} is the result obtained by repeating this procedure three times. In our method, it is observed that the magnitude of the error vector remains small regardless of the depth or 2D position in the image. In contrast, significantly larger error vectors are observed at closer distances for the checkerboard pattern. This phenomenon results from the measurement noise of control points, which increases at closer distances. For the remaining two methods based on circular patterns (i.e. conic-based and point-based), locally consistent large error vectors are observed. This local consistency indicates that these error vectors stem from the estimator's bias rather than measurement noise.

\subsection{Analytic solution of $\bs{T}^o_t$ and $\bs{T}^m_c$}
\label{Appendix:ax=xb}

\begin{align}
    \label{eq:motion1}
    \bs{X} &= \bs{T}^o_t,\\
    \label{eq:motion2}
    \bs{Y} &= \bs{T}^m_c,\\
    \label{eq:motion3}
    \bs{T}^m_{o_i}\bs{X} &= \bs{Y} \bs{T}^c_{t_i}
\end{align}
For obtaining optimal solution of $\bs{X}$ and $\bs{Y}$, we first decouple $\bs{X}$ and $\bs{Y}$ using \equref{eq:motion3} as
\begin{align}
    \label{eq:motion4}
    (\bs{T}^m_{o_j})^{-1}\bs{T}^m_{o_i} \bs{X}  &= \bs{X} (\bs{T}^c_{t_j})^{-1}\bs{T}^c_{t_i} & \text{for all ($i$, $j$) pair}\\
    \label{eq:motion5}
    \bs{T}^m_{o_j}(\bs{T}^m_{o_i})^{-1} \bs{Y}  &= \bs{Y} \bs{T}^c_{t_j}(\bs{T}^c_{t_i})^{-1} & \text{for all ($i$, $j$) pair}
\end{align}
Therefore, the remaining part is to solve the $\bs{A}_i\bs{X}=\bs{X}\bs{B}_i$ problem for $i=1\sim n$. According to the paper \cite{TRO-1994-frankP}, this problem has a closed-form solution.
\begin{align}
    \bs{X} &= \begin{bmatrix}
        \bs{R}_x & \bs{t}_{x}\\
        \bs{0}^{\top} & 1
    \end{bmatrix} = \begin{bmatrix}
        \exp([\bs{w}_x]) & \bs{t}_{x}\\
        \bs{0}^{\top} & 1
    \end{bmatrix},\\
    \bs{A}_i & = \begin{bmatrix}
        \exp([\bs{w}_{a_i}]) & \bs{t}_{a_i}\\
        \bs{0}^{\top} & 1
    \end{bmatrix} \hspace{0.3cm} \bs{B}_i  = \begin{bmatrix}
        \exp([\bs{w}_{b_i}]) & \bs{t}_{b_i}\\
        \bs{0}^{\top} & 1
    \end{bmatrix},\\
    \bs{P} &= \sum_i \bs{w}_{b_i}\bs{w}_{a_i}^{\top},\\
    \hat{\bs{R}}_x &= (\bs{M}^{\top}\bs{M})^{-1/2}\bs{M}^{\top},\\
    \bs{C} &= \begin{bmatrix}
        \bs{I} - \bs{R}_{a_1}\\
        \bs{I} - \bs{R}_{a_2}\\
        \cdots\\
        \bs{I} - \bs{R}_{a_n}\\
    \end{bmatrix}, \hspace{0.3cm} \bs{d} = \begin{bmatrix}
        \bs{t}_{a_1} - \hat{\bs{R}}_x\bs{t}_{b_1}\\
        \bs{t}_{a_2} - \hat{\bs{R}}_x\bs{t}_{b_2}\\
        \cdots\\
        \bs{t}_{a_n} - \hat{\bs{R}}_x\bs{t}_{b_n}\\
    \end{bmatrix},\\
    \hat{\bs{t}}_x &= (\bs{C}^{\top}\bs{C})^{-1}\bs{C}^{\top}\bs{d}.
\end{align}

\end{document}